\documentclass[conference]{IEEEtran}
\usepackage{times}
\usepackage{multicol}

\usepackage[numbers]{natbib}

\usepackage[utf8]{inputenc} %
\usepackage[T1]{fontenc}    %
\usepackage{breakcites}
\usepackage[bookmarks=true]{hyperref}       %
\usepackage{url}            %
\usepackage{booktabs}       %
\usepackage{amsfonts}       %
\usepackage{nicefrac}       %
\usepackage{microtype}      %

\usepackage{amsmath}
\usepackage{amssymb}
\usepackage{makecell}
\usepackage{todonotes}
\usepackage{graphicx}
\usepackage{cleveref}
\usepackage{siunitx}
\usepackage{subcaption}
\usepackage{wrapfig}
\usepackage{graphicx}
\usepackage{tikz}
\usepackage{algpseudocode,algorithm,algorithmicx}
\usepackage{mymath}
\usepackage{mlmacros}
\usepackage{geometry}

\usetikzlibrary{positioning,shapes,arrows}

\variables{z,x,u,m,y}
\probdists{p,q}
\probdists[dirac]{\delta}
\newcommand{\globalvar}{\boldsymbol{\omega}}
\newcommand{\genparam}{\boldsymbol{\theta}}
\newcommand{\infparam}{\boldsymbol{\phi}}
\newcommand{\lossobstacle}{\mathcal{L}_{\text{obstacle}}}
\newcommand*\Let[2]{\State #1 $\gets$ #2}

\newcommand{\showappendix}{1}       %
\newcommand{\ifappendix}[1]{\ifthenelse{\equal{\showappendix}{1}}{#1}{}}
\newcommand{\ifnotappendix}[1]{\ifthenelse{\equal{\showappendix}{0}}{#1}{}}

\crefname{appsec}{appendix}{appendices}

\begin{document}

\title{Approximate Bayesian inference in spatial environments}

\author{
\centering
\authorblockN{Atanas Mirchev\authorrefmark{2},
Baris Kayalibay\authorrefmark{2},
Maximilian Soelch\authorrefmark{2},
Patrick van der Smagt\authorrefmark{3} and
Justin Bayer\authorrefmark{2}
}
\authorblockA{\\
Machine Learning Research Lab, Volkswagen Group\\
80805, Munich, Germany \\
\authorrefmark{2}\{firstname.lastname\}@argmax.ai\\
\authorrefmark{3}smagt@argmax.ai}
}

\maketitle

\begin{abstract}
	Model-based approaches bear great promise for decision making of agents interacting with the physical world. 
In the context of spatial environments, different types of problems such as localisation, mapping, navigation or autonomous exploration are typically adressed with specialised methods, often relying on detailed knowledge of the system at hand. 
We express these tasks as probabilistic inference and planning under the umbrella of deep sequential generative models. 
Using the frameworks of variational inference and neural networks, our method inherits favourable properties such as flexibility, scalability and the ability to learn from data.
The method performs comparably to specialised state-of-the-art methodology in two distinct simulated environments.

\end{abstract}

\IEEEpeerreviewmaketitle

\section{INTRODUCTION}

Sequential decision making is a framework to represent the interaction of an agent with its environment: an observation of the world is presented to the agent, upon which the informed agent picks an action, which in turn alters the world's state.
One instance of interest are spatial environments such as mobile robots on a factory floor, autonomous cars, robot arms or unmanned aerial vehicles.
Various tasks are of interest in these scenarios.
Localisation or pose estimation considers the relation of the agent itself to the environment.
This is often combined with establishing a map of its surroundings and has been referred to as simultaneous localisation and mapping (SLAM) by the robotics community.
If the aim is to obtain that map efficiently, autonomous exploration is about devising trajectories that uncover the map with as little effort as possible.
Another goal is navigation, referring to the generation of plans that allow the agent to reach a pre-specified location.

\begin{figure}[t]
	\begin{center}
	\begin{subfigure}{0.33\linewidth}
		\tiny
		\includegraphics[height=3.2cm]{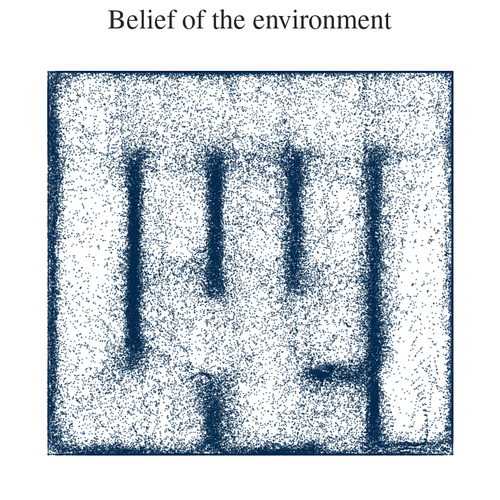}
		\caption{}
	\end{subfigure}\hfill
	\begin{subfigure}{0.33\linewidth}
		\tiny
		\includegraphics[height=3.2cm]{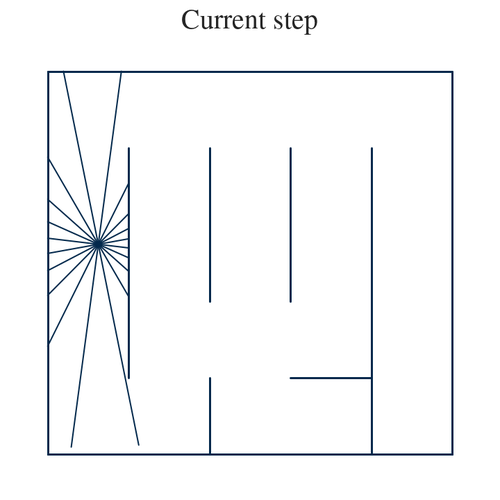}
		\caption{}
	\end{subfigure}\hfill
	\begin{subfigure}{0.33\linewidth}
		\tiny
		\includegraphics[height=3.2cm]{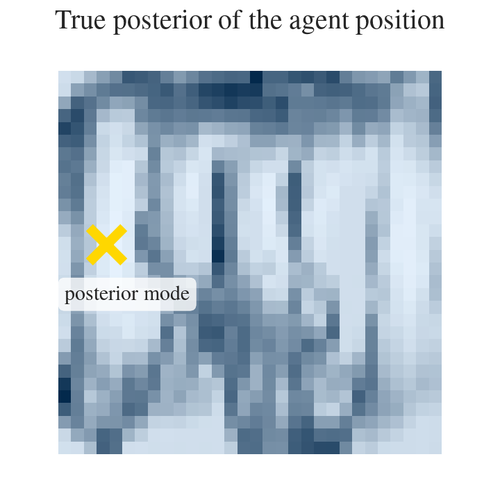}
		\caption{}
	\end{subfigure}\hfill
	\end{center}
	\caption{
		Example for the agent pose posterior in a LiDAR environment (pybox2d):
		(a) Scatter plot of observations predicted from the learned DVBF-LM map.
		(b) The current agent position and LiDAR readings.
		(c) Plot of the pose posterior over locations given the current observation.
		It is highly non-Gaussian and has several maxima.
	}
	\label{fig:intro}
\end{figure}

We set out to address these problems in a unified framework.
We augment a  non-linear state space model, the deep variational Bayes filter \cite{KarlSBS16} with a global latent variable representing a map (DVBF-LM).
We propose the necessary learning algorithms that enable end-to-end learning and make it possible to express the aforementioned tasks as either inference or planning in that model.
SLAM is performed by approximate, variational inference of the joint posterior over maps and pose trajectories.
We rely on a standard formulation as probabilistic inference in a graphical model.
For autonomous exploration, we optimise the expected information gain with respect to the control signals \cite{autoexploration, infogainrbpf}.
This is possible due to the probabilistic treatment of the map, where unexploredness is related to the remaining uncertainty in the respective region.
Navigation is implemented as planning in a discretisation of the learned model.

Our contributions are:
\begin{itemize}
	\item a deep non-linear state-space model with an explicit map component that can be estimated from data;
	\item methods of performing SLAM, autonomous exploration and navigation in said model;
	\item a solution to train stochastic recurrent models on a single, long, consecutive time series;
	\item a variational posterior formulation that copes with the complex joint posterior prevalent in SLAM (cf.\ \Cref{fig:intro}).
\end{itemize}
We validate the claims in a series of extensive experiments where we perform comparably to baselines tailored specifically to the respective settings.

\section{RELATED WORK}
The problem of concurrent estimation of an agent's pose and its surrounding has seen considerable attention in the last decades.
We refer the interested reader to the survey of \citet{Cadena2016Past}.
A contribution of \citet{Murphy1999Bayesian} is most similar to our approach: the map is a matrix-valued global latent variable inferred through Bayesian methods.
\begin{figure*}[t]
	\begin{minipage}{.28\textwidth}
		\centering
		\resizebox{0.9\columnwidth}{!}{\begin{tikzpicture}[
    -latex,
    auto,
    node distance=12em,
    on grid,
    semithick]

\definecolor{lightgray}{rgb}{0.8,0.8,0.8}

\tikzstyle{observed} = [circle, draw=black, fill=white, text=black, line width=.1em, minimum width=5em, minimum height=5em, font=\LARGE]
\tikzstyle{latent} = [circle, draw=black, fill=lightgray, text=black, line width=.1em, minimum width=5em, minimum height=5em, font=\LARGE]
\tikzstyle{invisible} = [draw=none, fill=none, text=black]
\tikzstyle{deterministic} = [diamond, draw=black, fill=lightgray, text=black, line width=.1em, minimum width=5em, minimum height=5em, font=\LARGE]

\node[observed] (x1) {$\bobs_t$};
\node[observed] (x2) [right=8em of x1] {$\bobs_{t+1}$};
\node[observed] (x3) [right=8em of x2] {$\bobs_{t+2}$};

\node[deterministic] (m1) [below=7em of x1] {$\bmap_t$};
\node[deterministic] (m2) [below=7em of x2] {$\bmap_{t+1}$};
\node[deterministic] (m3) [below=7em of x3] {$\bmap_{t+2}$};

\node[latent] (z1) [below=7em of m1] {$\bpose_t$};
\node[latent] (z2) [below=7em of m2] {$\bpose_{t+1}$};
\node[latent] (z3) [below=7em of m3] {$\bpose_{t+2}$};

\node[latent] (M) [below=7em of z2] {$\Map$};

\node[invisible] (past) [left=5em of z1] {$\dots$};
\node[invisible] (future) [right=5em of z3] {$\dots$};

\path[line width=.1em] (past) edge (z1);
\path[line width=.1em] (z1) edge (z2);
\path[line width=.1em] (z2) edge (z3);
\path[line width=.1em] (z3) edge (future);

\path[line width=.1em] (z1) edge (m1);
\path[line width=.1em] (z2) edge (m2);
\path[line width=.1em] (z3) edge (m3);

\path[line width=.1em] (m1) edge (x1);
\path[line width=.1em] (m2) edge (x2);
\path[line width=.1em] (m3) edge (x3);

\path[line width=.1em] (m1) edge (z2);
\path[line width=.1em] (m2) edge (z3);
\path[line width=.1em] (m3) edge (future);

\draw[line width=.1em] (M) to[out=140, in=-60] (m1);
\draw[line width=.1em] (M) to[out=130, in=215] (m2);
\draw[line width=.1em] (M) to[out=45, in=215] (m3);

\end{tikzpicture}\unskip}
		\captionof{figure}{Sequential graphical model with global map $\Map$ and local charts $\bmap_t$.}
		\label{fig:pgm}
	\end{minipage}%
	\hfill
	\begin{minipage}{.70\textwidth}
		\centering
		\includegraphics[width=\columnwidth]{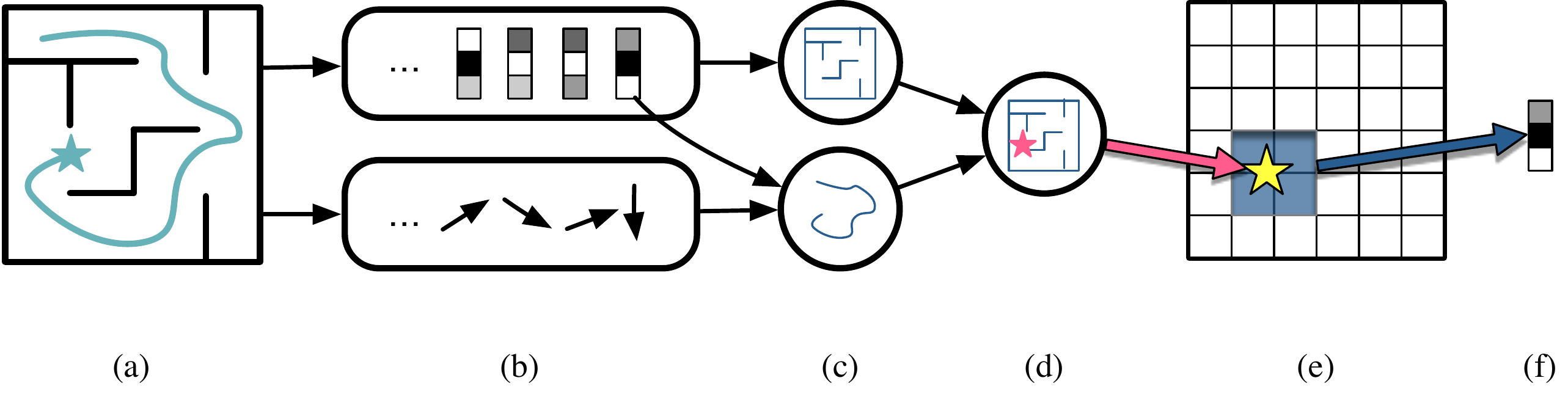}
		\captionof{figure}{
			\emph{Illustration of pose inference.}
			(a) An agent (teal star) traverses a maze,
			(b) collecting sensor readings (top) and control signals (bottom).
			(c) A belief of the map is formed from observations (top), a belief of the trajectory is formed from observations and controls (bottom).
			(d) The two beliefs are fused.
			(e) The map is indexed with a pose-based attention mechanism.
			(f) The attended region of the map is used to reconstruct the observation.
		}
		\label{fig:overview}
	\end{minipage}
\end{figure*}

Mapping and localisation has been adopted in the machine-learning community mostly to solve reinforcement-learning or visual-navigation problems \citep{Parisotto2017Neural,Savinov2018Semi}.
\citet{Fraccaro2018Generative} proposed a generative model for spatial environments.
While their approach is similar to ours, their focus was primarily on simulator performance over long time spans.
Further, an external memory is used which does not directly represent a random variable as part of a graphical model.

Early discussions of the importance of exploration of model parameters can be found in \cite{curiosity,thrunexplore}, and information-theoretic methods for spatial exploration can be traced back to \cite{oldexplore,oldexplore2}.
Our work follows the framework of \emph{curiosity-driven exploration} \citep{curiosity}.
In the spatial environment context, it represents an instance of active SLAM \citep{autoexploration,Cadena2016Past}.

A large body of recent spatial exploration methods \cite{infogainrbpf,gpexploration,octomapsexplore,tensorfields} is driven by information theory, but assumes particular discretisations of the state space (occupancy grids) or skeletonisations of the possible action paths.
Variational information maximizing exploration (VIME) \cite{vime} is closely related to our method, but imposes an intentional Gaussian constraint for tractability and mutual information is estimated for one step $t$ at a time.
Our method has no such assumptions.

\section{DEEP VARIATIONAL BAYES FILTER WITH A LATENT MAP}
\label{sec:methods}
Our aim is to provide a unified model that covers all of the aforementioned tasks.
We defer the discussion of their exact implementations to \Cref{sec:slam}, \Cref{sec:navigation} and \Cref{sec:exploration}.
Here we will focus on the underlying probabilistic generative model.
We step on the solid foundation of state space models to represent sequential agent interactions:
\setlength{\abovedisplayskip}{6pt}
\setlength{\belowdisplayskip}{6pt}
\eq{
	&\p{\bx\Ts, \bz\Ts}{\bu\Tsm} =\\
	&~\p{\bz_1}
	\prod_{t=2}^T \p{\bz_t}{\bz_{t-1}, \bu_{t-1}}
	\prod_{t=1}^T \p{\bx_t}{\bz_t},
}
where 
\medbreak
\setlength\tabcolsep{6.pt} %
\begin{tabular}{ll}
	$\bobs_{1:T} \in \RR^{T \times D_x}$&is a sequence of observations,\\
	$\bpose_{1:T} \in \RR^{T \times D_z}$&is a sequence of poses, and\\
	$\bcontrol_{1:T-1} \in \RR^{T-1 \times D_u}$&is a sequence of control inputs. 
\end{tabular}
\medbreak
In terms of spatial environments, the \emph{transition model} $\p{\bz_t}{\bz_{t-1}, \bu_{t-1}}$ represents the dynamics and the \emph{emission model} $\p{\bx_t}{\bz_t}$ simulates the formation of observations.

The traditional state space model above is a powerful tool for the analysis of stochastic dynamical systems \cite{KarlSBS16}.
In order to unlock further inferences specific to spatial environments, more transparency is required.
We thus introduce additional structure to provide the necessary entry points.
First, we identify a part of the latent space with the environment itself.
To that end we extend the traditional graphical model, incorporating a global map latent variable $\Map\sim\p{\Map}$.
We introduce a latent middle layer of local charts $\bmap_{t}\sim\pp{\bmap_t}{\bpose_t, \Map}$.
Intuitively, the chart $\bmap_t$ represents the currently relevant attended region of the map.
It shapes the transition of poses over time, $\pp{\bpose_{t+1}}{\bpose_t, \bmap_{t}, \bcontrol_{t}}$.
The observation emission model operates solely on these charts, $\bobs_t \sim \pp{\bobs_t}{\bmap_t}$.
In total, this yields the graphical model (cf.\ \Cref{fig:pgm})
\eq{
	&\pp{\bobs_{1:T},\bpose_{1:T},\bmap_{1:T},\Map}{\bcontrol_{1:T-1}}= \\
	&\,\qquad \p{\Map} \rho(\bpose_1)\\
	&\qquad \prod_{t=1}^T \pp{\bobs_t}{\bmap_t} \pp{\bmap_t}{\bpose_t, \Map} \\
	&\qquad \prod_{t=1}^{T-1}\pp{\bpose_{t+1}}{\bpose_t, \bmap_{t}, \bcontrol_{t}}.
}
A common assumption is that the factors are governed by a set of parameters $\genpars$, i.e.\ $\rho_{\genpars_I}(\bpose_1)$, $\ppu{\bpose_{t+1}}{\bpose_t, \bcontrol_t, \bmap_t}{\genpars_T}$, $\ppu{\bobs_{t}}{\bmap_t}{\genpars_E}$.
We will leave out the dependency for notational brevity in the remainder of this work.
Further, in the conducted experiments we assume the transition parameters $\genparam_T$ and the initial state distribution are learned \emph{a priori} or engineered.

\subsection{Approximation via Variational Inference}
\label{sec:approximation}
Exact inference in such models is typically intractable.
We obtain variational approximations $\q{\Map}$ and $\qq{\bpose_{1:T}}{\bobs_{1:T}, \bcontrol_{1:T-1}, \Map}$ of the corresponding posteriors, collecting their learnable variational parameters in $\varpars$, where we rely on Bayes by backprop \citep{blundel2015} for the former and on SGVB \citep{kingma2014auto} for the latter.
The negative \emph{evidence lower bound} (ELBO) is given as 
\eq{
\elbo = ~& \underbrace{\expcc{-\log \pp{\bobs_{1:T}}{\bpose_{1:T}, \Map}}{q}}_{=: \recloss} + \\
& \underbrace{\kl{\q{\Map}}{\p{\Map}}}_{=: \mappenalty} + \\
& \underbrace{\expcc{\kl{\qq{\bpose_{1:T}}{\bobs_{1:T}, \Map}}{\pp{\bpose_{1:T}}{\Map}}}{q}}_{=: \posepenalty}. \label{eq:elbo} \numberthis
}
The conditioning on $\bcontrol_{1:T-1}$ is dropped for brevity.
We call $\recloss$ the \emph{reconstruction loss}, $\posepenalty$ the \emph{pose KL penalty} and $\mappenalty$ the \emph{map KL penalty}.
Inference of poses and the map then comes down to the minimisation of \Cref{eq:elbo} with respect to $\varpars$.

\subsection{Implementation of the Generative Model}
\label{sec:implementation}
In general, the geometric properties of the environment that need to be represented will determine the particular form of the map and the associated attention model.
For the purposes of this work we follow \cite{Murphy1999Bayesian}, defining the map $\Map$ to be a finite grid of width $w$ and height $h$.
Each grid cell $\Map_{ij}$ is a real-valued vector of dimensionality $D_m$, i.e.\  $\Map \in \RR^{w \times h \times D_m}$.
As prior for such a latent map cell we use a standard normal, $\Map_{ij} \sim \mcN(\mathbf{0}, \mathbf{1}).$
Extracting \emph{local charts} $\bmap_t$ from the map is done through a convex combination of the memory cells:
\eq{
	\bmap_t =f_m(\bpose_t, \Map) =\sum_{i, j} \attention(\bpose_t)_{ij} \Map_{ij}.
}
The result is then a point mass:
\eq{
	\pp{\bmap_t}{\bpose_t, \Map, \genpars_M} \propto \mathbb{I}\left [ \bmap_t = f_m(\bpose_t, \Map)\right ].
}
In this implementation, we choose $\alpha$ to be a bilinear interpolation kernel, combining four cells at a time.

The \emph{emission model} and \emph{transition model} are conditional Gaussian distributions with fixed diagonal covariances. The respective means are given by neural networks parameterised by $\genpars_E$ and $\genpars_T$:
\eq{
       \pp{\bobs_t}{\bmap_t} =&~ \mcN \left(\bmu_E( \bmap_t), \text{diag}(\bsigma_E^2)\right), \\
    \pp{\bpose_{t+1}}{\bpose_t, \bcontrol_t, \bmap_t} =&~ \mathcal{N} \left(\bmu_T(\bpose_t, \bcontrol_t, \bmap_t), \sigma_T^2 \mathbf{1} \right).
}

\subsection{Design of the Variational Posterior}
\label{sec:design}
Inference of poses is done through a variational approximation of the true posterior $\qq{\bpose_t}{\bobs_{1:T}} \approx \pp{\bpose_t}{\bobs_{1:T}},$ where we left out the control signals $\bcontrol_{1:T-1}$ for brevity and will do so for the remainder of this section.
The global variable $\Map$ poses an atypical challenge for stochastic recurrent models trained with amortised variational inference, for which an intuitive explanation is as follows.
Consider the true posterior, which has to account for all possible maps:
\eq{
\pp{\bpose_{t}}{\bobs_{1:T}} =
\int 
\pp{\bpose_{t}}{\Map, \bobs_{1:T}}
\pp{\Map}{\bobs_{1:T}}
d\Map.
}
Any parameterised variational approximation $\qq{\bpose_t}{\bobs_{1:T}}$ will have to implement its own belief of the map implicitly.
During training, this will prove difficult as it has to track the current belief of the generative model to conform to it, as it essentially implements its inverse.
The task of the inference model can be substantially eased by informing it of the current belief of the map $\q{\Map}$ explicitly.
We choose to do so by implementing $q$ as a bootstrap particle filter \cite{bootstrapfilter} with the particle forwarding distribution from \Cref{sec:approximate} as a proposal distribution:
\eq{
       \qq{\bpose_{1:T}}{\bobs_{1:T}, \Map} =&~ \prod_{t=1}^T \qq{\bpose_t}{\bobs_{1:t}, \Map}, \\
       \qq{\bpose_t}{\bobs_{1:t}, \Map} \propto&~ \expcc{\sum_{k=1}^K \hat\impweight_i \mathbb{I}(\bpose_t = \bpose_t^{(k)}) }{\bpose_t\supidx{k} \sim \hatq{\bpose_t}}, \\
    \hat\impweight_k =&~ \frac{\impweight_k}{\sum_j \impweight_j}, \\
    \impweight_k =&~ \frac{\pp{\bobs_t}{\bpose_t\supidx{k}, \Map}\p{\bpose_t\supidx{k}}}{\hatq{\bpose_t\supidx{k}}}.
}

This has two immediate consequences.
First, the variational posterior used does not have any parameters and is hence not optimised directly.
Second, the true posterior is recovered for $K \rightarrow \infty$.
But most importantly, the importance weights explicitly reflect the map (sampled from an outer expectation over $\q{\Map}$ in \Cref{eq:elbo}) and the proposals in conflict with it will be sorted out in a natural manner as they have lower weights.

The variational approximation of the posterior map $\q{\Map}$ was chosen to follow a mean-field approach with a factorised Gaussian $\q{\Map} = \prod_i \prod_j \mcN(\bmu_{\Map_{ij}}, \bsigma^2_{\Map_{ij}})$, with variational parameters $\bmu_{\Map_{ij}}, \bsigma^2_{\Map_{ij}} \in \varpars$.

\subsection{Faster Training with Mini Batches}
\label{sec:faster}
In practice, inference is typically performed on very long, continuous streams of data recorded from a moving agent.
Evaluating the ELBO for the whole trajectory at once proves prohibitive for learning or is downright impossible due to memory limitations.
We therefore seek to relax the optimisation while still respecting the underlying model.
\subsubsection{Decomposing the Loss into a Sum over Time Steps}
\label{sec:decomposing}
Under the Markov assumptions, the evidence lower bound from \Cref{eq:elbo} can be written as a sum over time steps:
\eq{
       \elbo &= \recloss + \posepenalty + \mappenalty = \expcc{\sum_{t=1}^T \recloss_t + \posepenalty_t + \mappenalty_t}{q} \numberthis \label{eq:decomposition}
}
with, leaving out the control signals $\bcontrol_{1:T-1}$ for brevity:
\eq{
       \recloss_t &= -\log \pp{\bobs_t}{\bpose_t, \Map}, \quad &\recloss &=  \expcc{\sum_{t=1}^T \recloss_t}{q}, \\
	\posepenalty_t &= \log \frac{\q{\bpose_t}{\bobs_{1:t}, \Map}}{\pp{\bpose_t}{\bpose_{t-1}, \Map}}, \quad &\posepenalty &= \expcc{\sum_{t=1}^T \posepenalty_t}{q}, \\
	\mappenalty_t &= \frac{1}{T} \log \frac{\q{\Map}}{\p{\Map}}, \quad &\mappenalty &= \expcc{\sum_{t=1}^T \mappenalty_t}{q}.
}
Following \cite{blundel2015}, we distribute the contribution of the map KL penalty term over different time steps, reflected in $\mappenalty_t$.
We denote the overall loss at time step $t$ as $\loss_t$. 

If the loss function is a sum over independent terms, a gradient estimator using only a subset of those terms will be unbiased.
Unfortunately the terms for each time step in \Cref{eq:decomposition} are not independent, which requires ancestral sampling from the whole Markov chain.

\subsubsection{Approximate Asynchronous Particle Representation}
\label{sec:approximate}
To overcome this issue we maintain sets of $N$ particles $\bparticle\supidx{n}_t, n=1, \dots, N; t=1, \dots, T$ that cache samples for each step of the variational posterior over poses $\q{\bpose_{1:T}}{\bobs_{1:T}, \Map} = \prod_{t=1}^T \q{\bpose_t}{\bobs_{1:t}, \Map}$ during training.

We define an estimator of the gradients from the complete ELBO, akin to stochastic gradient descent.
The time steps we wish to use for gradient estimation are gathered in a minibatch $\minibatch$.
We then approximate the loss given in \Cref{eq:decomposition} via
\eq{
	\tilde\loss = \frac{T}{|\minibatch|} \sum_{t \in \minibatch} \expcc{\expcc{\recloss_t + \posepenalty_t + \mappenalty_t}{\bpose_{t} \sim \tilde{q}}}{\Map \sim q}, \numberthis \label{eq:subsampled}
}
where $\tilde q(\bpose_{1:T}) = \prod_{t=1}^T \tilde q(\bpose_t)$ is an approximation of $\q{\bpose_{1:T}}{\bobs_{1:T}, \Map}$ based on the cached particles that allows more efficient sampling of $\bpose_t$.
In particular, every $\tilde q(\bpose_t)$ is importance-resampled from an underlying proposal distribution $\hat q(\bpose_t)$, which in turn is based on the particles $\bparticle\supidx{n}_t, n=1, \dots, N$.
In this work, $\hat q(\bpose_t)$ is represented as a Normal random variable with moments matched from the set of $N$ particles for time step $t$: $\hat q(\bpose_t) = \mathcal{N}(\bmu_{\bparticle_t}, \bsigma_{\bparticle_t}^2)$.
The mean $\bmu_{\bparticle_t} = \frac{1}{N}\sum_{n=1}^N \bparticle_t\supidx{n}$ and variance $\bsigma_{\bparticle_t}^2 = \frac{1}{N}\sum_{n=1}^N (\bmu_{\bparticle_t} - \bparticle_t\supidx{n})^2$ are the empirical mean and variance of the particles respectively.
The approximating particle sets are updated during gradient estimation: for any training iteration with $t \in \minibatch$, we can update the particles at following time steps $t+k, k = 1, \dots$: %
\eq{
	\bparticle_{t+k}\supidx{n} \sim \tilde q(\bpose_t) \prod_{i=1}^k \pp{\bpose_{t+i}}{\bpose_{t+i-1}, \Map},
}
effectively performing importance resampling and moving particles forward through the transition model, refreshing the approximation $\hat q(\bpose_{t+k})$.
This leads to an asynchronous procedure: expectations in \Cref{eq:decomposition} w.r.t. the approximate posterior over agent poses are implemented through particles stemming from previous training iterations, potentially biasing the gradients.
This bias can be controlled with small parameter updates (i.e.\ $\varpars\supidx{i+1} \approx \varpars\supidx{i}$), since we can then expect the expectations to be close as well.
In practice, we will choose chunks of consecutive time steps to be the elements of mini batches, requiring to only update the particles at the beginning of each such chunk.

\section{DVBF-LM AS A METHOD FOR SLAM}\label{sec:slam}
We first investigate the capabilities of DVBF-LM as a solution to SLAM.
Our model is evaluated in two simulated, precisely controlled environments---a 2D environment with laser range finder observations and VizDoom \citep{Kempka2016ViZDoom}.
A detailed description of each, along with additional information regarding the experimental setup can be found in the supplementary material \footnote{Available at: \url{https://arxiv.org/abs/1805.07206}.}.
\ifappendix{A detailed description of each can be found in \Cref{sec:simulator}.}

We randomised seven distinct 2D maze patterns and replicated them in both environments.
Each maze was traversed multiple times by two human operators to collect data\ifappendix{ (cf. \Cref{sec:simulator})}.
For both environments, the transition model $\pp{\bpose_{t+1}}{\bpose_t, \bcontrol_t, \bmap_t}$ is pretrained on a first maze that is not considered during evaluation.
Performing SLAM then consists of approximating the posterior of the poses and the map $\pp{\bpose_{1:t}, \Map}{\bobs_{1:t}, \bcontrol_{1:t-1}}$ for a single traversal through the optimisation of \Cref{eq:elbo} with respect to the variational posteriors $\q{\bpose_{1:T}}$ and $\q{\Map}$.

We consider two cases, \emph{offline} and \emph{online SLAM}.
In the first case we optimise for all time steps at once, whereas in the second case we sweep $t=1, \dots, T$ to obtain time-step-wise estimates $\q{\bpose_{1:t}}$.

All distances in the following experiments are unit-less, the width and height of the considered mazes were set to $1.0$.

\subsection{Pybox2d Environment}
For this environment we implemented our own 2D simulator using pybox2d\ifappendix{ (cf. \Cref{sec:simulator} for simulation details)}, in which the agent's sensors are laser range finders (LiDAR readings).
\paragraph{Improving Path Integration}
The aim of this set of experiments is to test whether the proposed approximation of the graphical model improves upon direct path integration based on the pretrained transition $\pp{\bpose_{t+1}}{\bpose_t, \bcontrol_t, \bmap_t}$ only.
For both online and offline SLAM, using a map clearly outperforms the path integration baseline in terms of localisation error: $0.03 \pm 0.02$ and $0.04 \pm 0.02$ at time step 3000 for online and offline SLAM respectively.
At this time step, the motion model has diverged for most of the sequences with an average error of $0.14 \pm 0.1$.
Most notably, the use of a map practically eliminates drift: after 3000 steps, a relative error of less than $20{,}000^{-1}$, effectively zero, is obtained.
This shows that our method stabilises the motion model and keeps the location estimate from diverging. 
We illustrate the findings in \Cref{fig:results}.

\begin{figure}[t]
	\begin{center}
        \includegraphics[width=\linewidth,trim={23cm 0 0 0},clip]{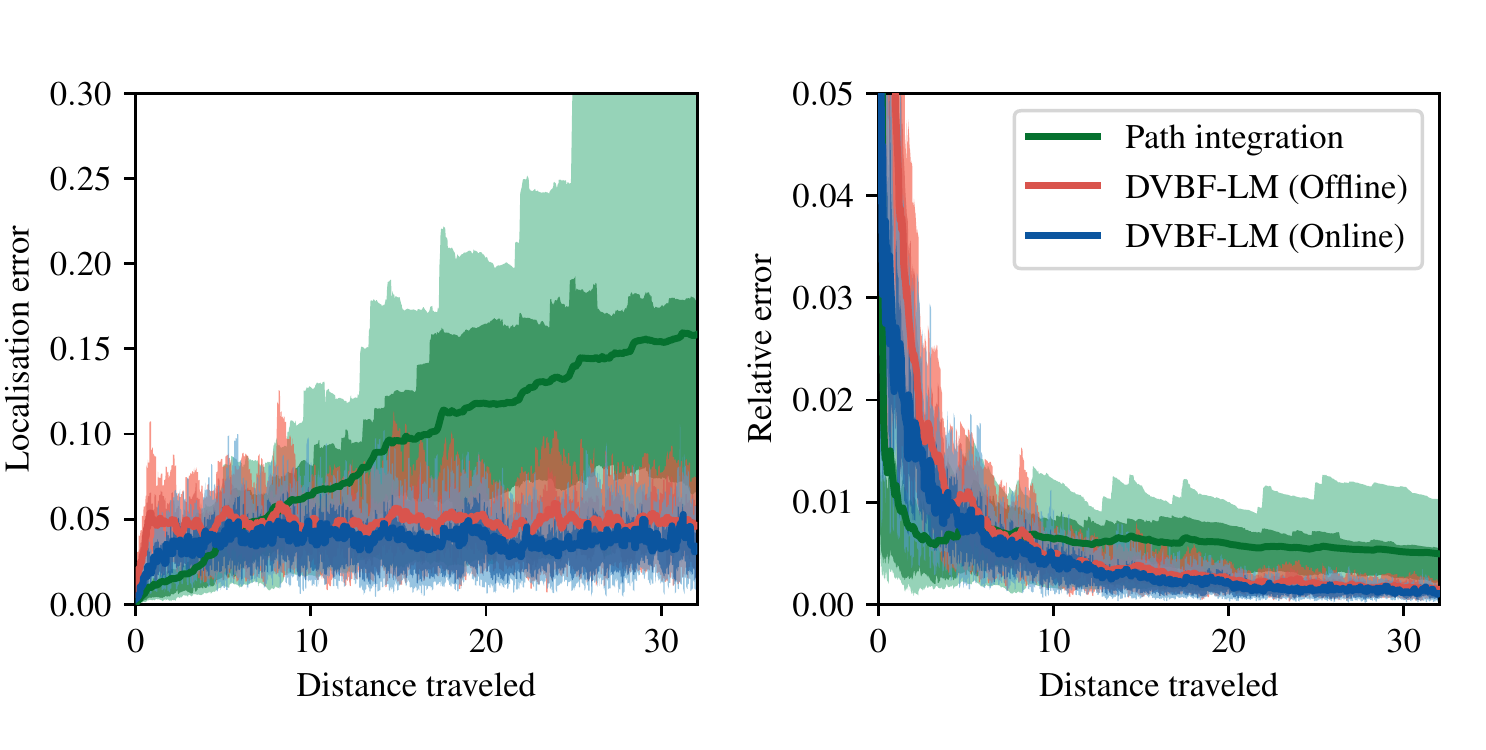}
	\end{center}
	\caption{
		We compare the online and offline DVBF-LM localisation error to path integration for all 24 test traversals accross 6 mazes in the pybox2d environment.
		The plots show aggregate results, shaded regions contain 50\% and 80\% of the traversals.
		The second plot shows the localisation error relative to the distance travelled.
	}
	\label{fig:results}
\end{figure}

\paragraph{Comparison to Cartographer}
Next we compare DVBF-LM's online localisation performance to that of Google's Cartographer \citep{Hess2016Real}, which we consider a representative baseline model for 2D LiDAR SLAM.
Cartographer is a realtime SLAM system which operates on laser range finder data and is capable of detecting loop closures.
In addition to the LiDAR observations collected from pybox2d, we provided Cartographer with the angular velocity of the agent at every time step in the form of IMU readings.
In order to improve upon the default Cartographer configuration and tune it to our setup, we performed a hyperparameter search over more than 40 of Cartographer's hyperparameters, with 6000 trials on a held out trajectory of 1000 steps.
Both Cartographer and DVBF-LM manage to eliminate drift, with respective errors of $0.05 \pm 0.04$ and $0.03 \pm 0.02$ at time step 3000. 
The proposed graphical model approximation leads to localisation performance that is consistently on par in quality to that of Cartographer.
The results from the comparison are depicted in \Cref{fig:cartographer}.

\subsection{VizDoom Environment}

The VizDoom experiments take place in the same set of mazes as pybox2d.
Observations are now two-dimensional images taken from the perspective of the agent.
\ifappendix{Details on the simulator design can be found in \Cref{sec:vizdoomenv}.}

For the VizDoom environment we only investigate offline SLAM performance.
The experiment procedure was identical to the pybox2d counterpart.
All model components apart from the emission model are kept the same.
The latter is modified to better fit visual observations.
\ifappendix{See \Cref{sec:vizdoomemitdetails}.}

We summarise quantitative localisation results in \Cref{fig:vizdoom}.
The method performs on a similar level as in the laser scan-based environment: localisation error is $0.04 \pm 0.03$ after $5000$ time steps.
We can see that DVBF-LM is capable of correcting the drift resulting from path integration (an error of $0.11 \pm 0.05$).
The final relative error is $0.06\% \pm 0.06\%$ of the trajectory length.
The results indicate that our method is able to perform accurate localisation when applied to different observation modalities in an offline fashion by adapting only the architecture of the emission model.

\begin{figure}[t]
	\begin{subfigure}[t]{\linewidth}
		\centering
		\includegraphics[width=\linewidth,trim={19.3cm 0 0 0},clip]{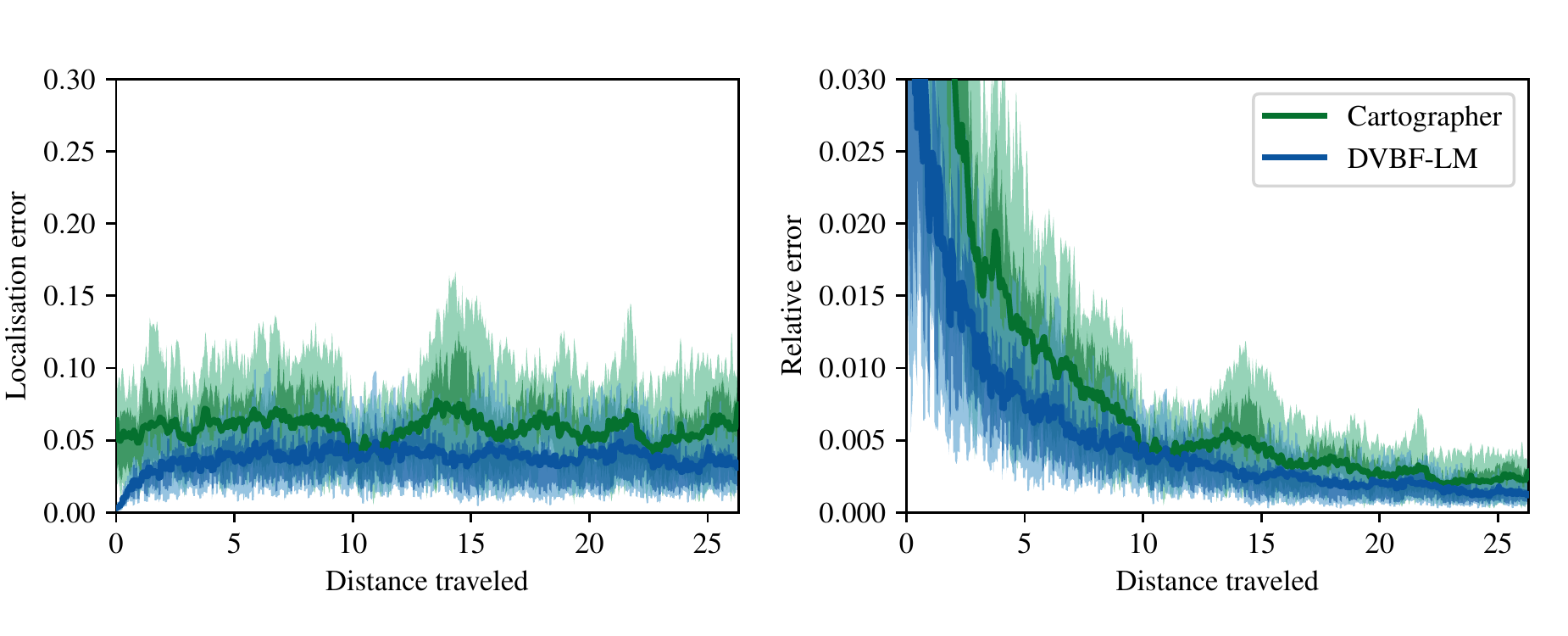}
		\caption{Pybox2d}
		\label{fig:cartographer}
	\end{subfigure}

	\begin{subfigure}[t]{\linewidth}
		\centering
		\includegraphics[width=\linewidth,trim={19.3cm 0 0 0},clip]{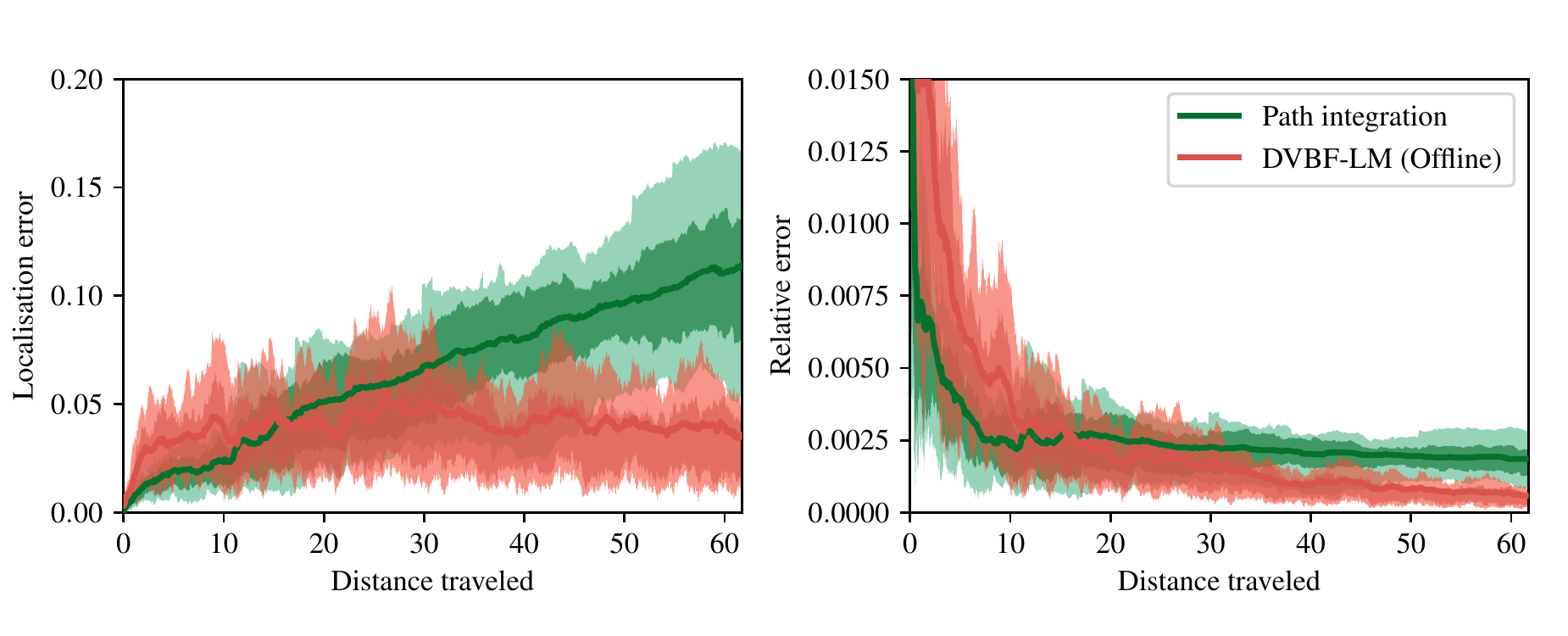}
		\caption{VizDoom}
		\label{fig:vizdoom}
	\end{subfigure}
	\caption{
		(\subref{fig:cartographer}) Online DVBF-LM localisation error compared to that of Cartographer on test mazes in the pybox2d environment. We omit the offline DVBF-LM results from \Cref{fig:results} for the sake of legibility.
		(\subref{fig:vizdoom}) Offline DVBF-LM localisation error compared to path integration on test maze traversals in the VizDoom environment.
		The plots show aggregate results, shaded regions contain 50\% and 80\% of the traversals.
		The plots on the right show the localisation error relative to the distance travelled.
	}
\end{figure}

\section{NAVIGATION IN LEARNED ENVIRONMENTS}
\label{sec:navigation}
One incentive for learning a generative model of a spatial environment is that such a model can be used to plan interactions with that environment.
As an example, we use DVBF-LM as a black-box environment simulator and provide its predictions to a classical path planning algorithm to solve navigation tasks.

\newcommand{\astar}{$A^*$}
\subsection{Latent Hybrid-\astar}
\label{sub:astar}
The latent map $\Map$ of DVBF-LM conserves the Euclidean geometry of the true environment through the inductive bias of the pretrained transition.
Hence, we are able to use the \emph{hybrid-\astar} algorithm \citep{dolgov_path_2010} for path planning.
The goal of hybrid-\astar\ is to find a path from a \emph{continuous} starting pose to a \emph{continuous} target pose.
This is done by discretising the search space into a grid of $N$ cells $\{\mathbf{c}_n\}_{n \in [N]}$, which the conventional \astar-search \cite{astar} can operate on.

To obtain smooth navigation trajectories, every discrete state $\mathbf{c}_n$ is associated with a continuous state $\mathbf{z}_n$---the agent's state when that cell was explored for the first time.
New cells $\mathbf{c}_{n+1}$ are explored by picking random sequences of controls $\mathbf{u}_{1:K}^n$ and predicting a following continuous state by applying the controls to the current $\mathbf{z}_n$.
The successor states are found by applying the transition model of DVBF-LM and picking the mean of the resulting Gaussian distribution.
When the target state is found, we backtrack to obtain a consistent sequence of controls $\mathbf{u}_{1:T}$, which can be executed by the agent to reach the target.

\begin{figure*}[t]
   \centering
   \hfill
	\begin{subfigure}{0.25\textwidth}
	   	\centering
		\small
	   	Planning in belief space
	   	\includegraphics[width=1.0\linewidth]{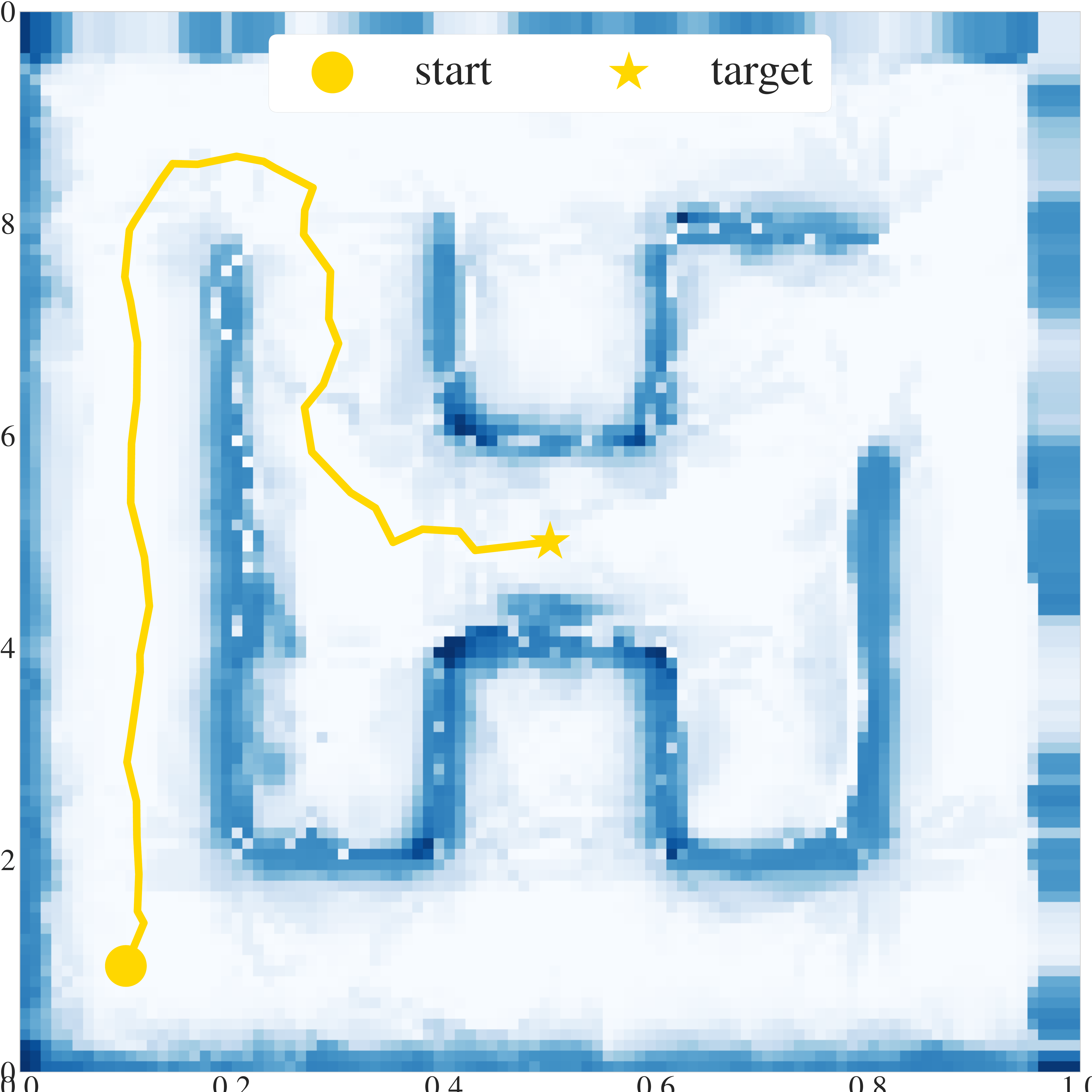}
	   	\caption{}
	   	\label{fig:traj_pred}
	\end{subfigure}\hfill
	\begin{subfigure}{0.25\textwidth}
	   	\centering
		\small
	   	Execution in real world
	   	\includegraphics[width=1.0\linewidth]{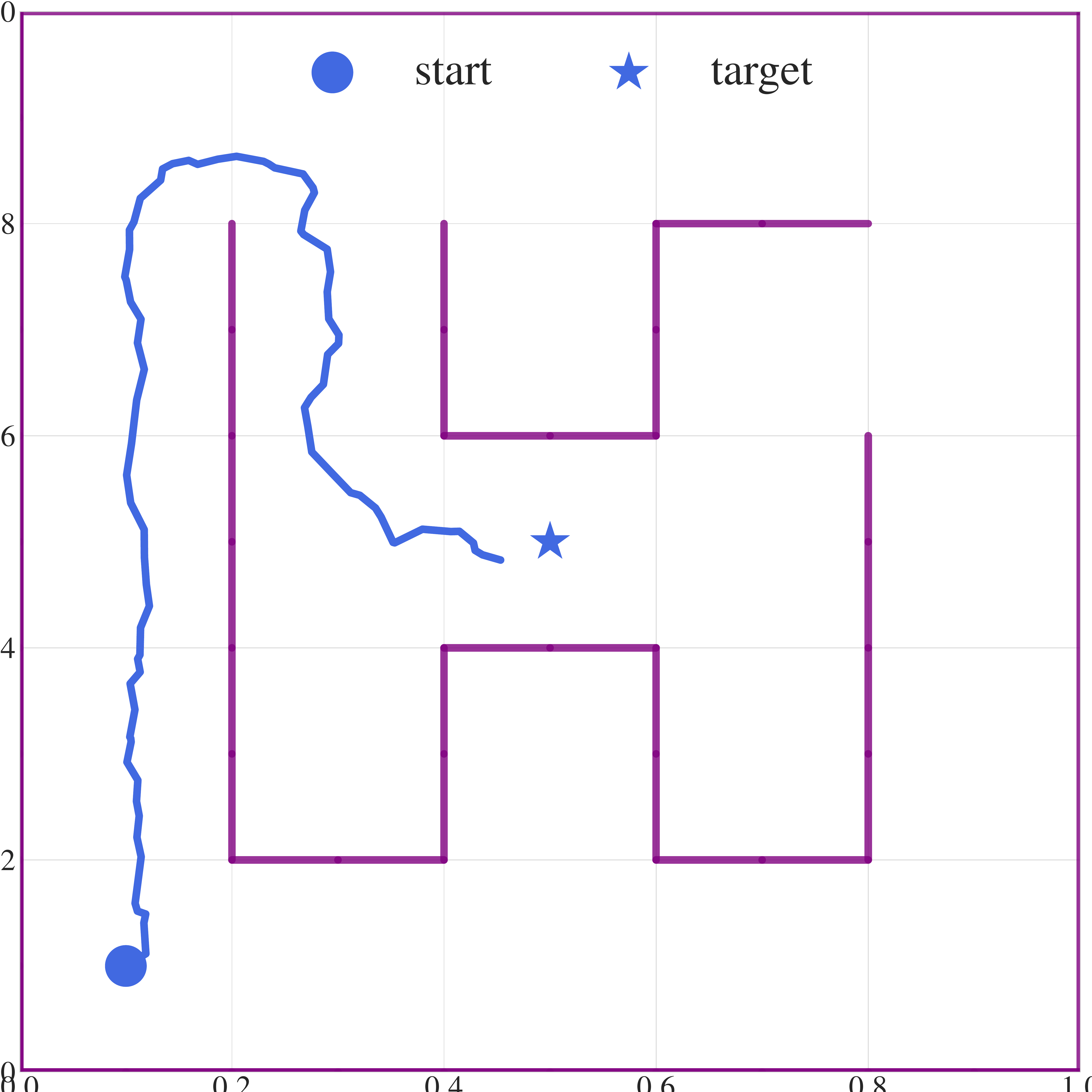}
	   	\caption{}
	   	\label{fig:traj_truth}
	\end{subfigure}\hfill
	\begin{subfigure}{0.25\textwidth}
		\centering
		\small
		Aggregate navigation success
		\includegraphics[width=1.0\linewidth]{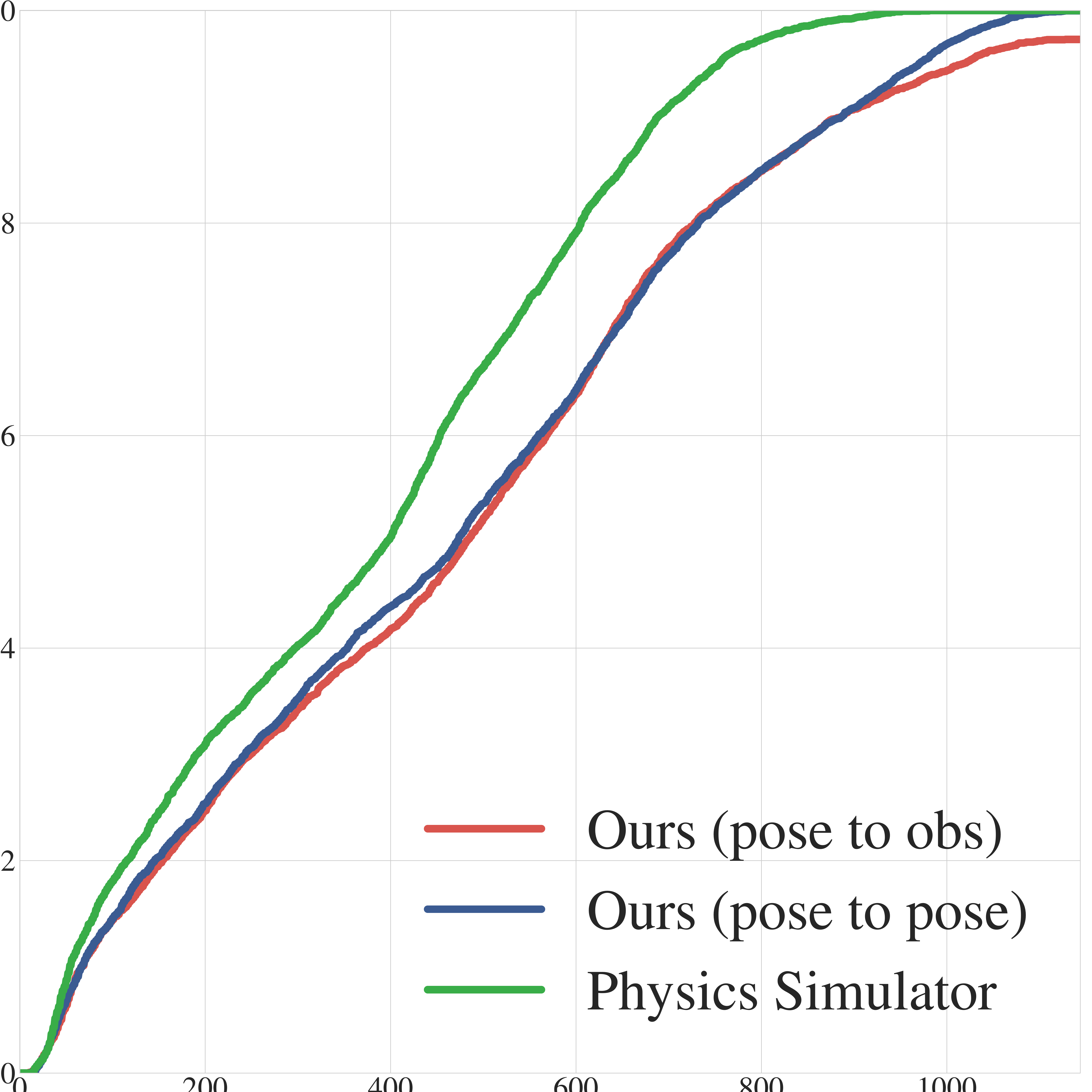}
		\caption{}
		\label{fig:navigationaggr}
	\end{subfigure}\hfill
   \hfill
   \label{fig:navresults}
   \caption{
  	(\subref{fig:traj_pred}) Navigation plan constructed by using the belief space of the learned environment. 
  	(\subref{fig:traj_truth}) Executed trajectory in the actual simulator. 
   	(\subref{fig:navigationaggr}) Ratio of successful navigation attempts over the number of simulation steps in \astar. In green the upper bound, \astar~with access to the ground truth map and transition.
}
\end{figure*}

One issue with using an approximate transition model is that collisions with obstacles cannot always be modelled accurately.
The problem is exacerbated by the fact that shortest paths in spatial environments tend to stay close to obstacles.
In order to alleviate the negative effect of these two factors on navigation success, we introduce a safety term which penalises closeness to obstacles: $\mathcal{L}_{\mathrm{safe}} = \sum_{i}s(l_i)$.
Here, $l_i$ is the reading of the $i$-th range sensor of the agent, as predicted by DVBF-LM based on the learned map $\q{\Map}$, and $s(\cdot)$ is a sigmoidal function mirrored along the $y$-axis.
\ifappendix{For details regarding the parameterisation of this function, see \Cref{sub:safety}.}

We add $\mathcal{L}_{\mathrm{safe}}$ to the travel distance when calculating edge weights.
The penalty term assumes that we have access to depth readings, which is true in the case of laser scan-based settings but not when the agent only has access to visual observations.
Generalising this path planning framework to visual observations will be part of future work.

\subsection{Results}
\label{sub:navresults}
\label{sec:experiments}
In our experiments, we verify that the model described in \Cref{sec:methods} can be applied to navigation.

\paragraph{Generative Model Learning}
Before DVBF-LM can be used for planning, the emission model and the map $\Map$ must first be learned.
Here, we use the same models that were acquired as part of the SLAM experiments.
For the transition we use an engineered model that allows the agent to move forward along its heading as long as the predicted LiDAR reading in that direction is greater than the desired step length.
All navigation tasks are performed in the same set of mazes introduced earlier.

\paragraph{Pose-to-Pose Navigation with Hybrid-\astar} For each maze, we exhaustively pick pairs of starting and target pose from a $5\times5$ grid over the map.
For each pair of poses, we apply the hybrid-\astar\ as described in \Cref{sub:astar} to plan a trajectory.
The obtained controls are then executed in the true physics simulator.
The navigation task is considered successful if the agent lands in a proximity of $0.05$ or less from the target pose.
As a baseline, we consider the navigation performance when the planning algorithm is executed directly in the physics simulator, with access to the true transition and emission.
This allows us to assess the drop in performance resulting from approximating the true environment with DVBF-LM.
\Cref{fig:navigationaggr} shows the results of the evaluation.
Planning based on our generative model comes very close in terms of navigation efficiency to planning in the simulator, affirming the usability of the learned environment maps for navigation tasks. Furthermore, all planned trajectories are successful in reaching the target.

\paragraph{Pose-to-Observation Navigation with Hybrid-\astar} In this scenario, the observation targets are sensor readings from the environment simulator. 
The corresponding starting poses are the same as in the previous case.
Before we can apply the algorithm from \ref{sub:astar} the observation targets must first be mapped to pose targets.
To that end, we train a separate variational auto-encoder (VAE, \cite{kingma2014auto}) on the same data used for learning a map of the environment.
We set the generative part of the VAE to the emission model of DVBF-LM, we condition on the learned map $\q{\Map}$, and freeze the map and emission parameters.
Thus, we obtain an approximation $\qq{\bpose}{\bobs}$ of the posterior over poses $\pp{\bpose}{\bobs}$ that conforms to the learned spatial map.
This is done once for each of the six mazes.
The obtained approximation can be reused for multiple navigation tasks in the given environment.
The rest of the evaluation proceeds analogously to the \emph{pose-to-pose} case, using the mode of the approximate posterior, $\bpose^* = \arg \max_{\bpose} \qq{\bpose}{\bobs}$, as a target.
Figure \ref{fig:navigationaggr} illustrates the results of the evaluation.
Performance is very similar to the \emph{pose-to-pose} case, with less than $2\%$ of all trajectories failing to reach their actual target.
The slight drop in performance can be attributed to \emph{perceptual aliasing}---ambiguity in the pose given an observation---that is typical for spatial environments.

\begin{figure*}[t]
\begin{minipage}{.35\textwidth}
	\begin{subfigure}[t]{0.49\linewidth}
		\centering
		\small
		Candidate trajectories \\
		\includegraphics[width=\linewidth]{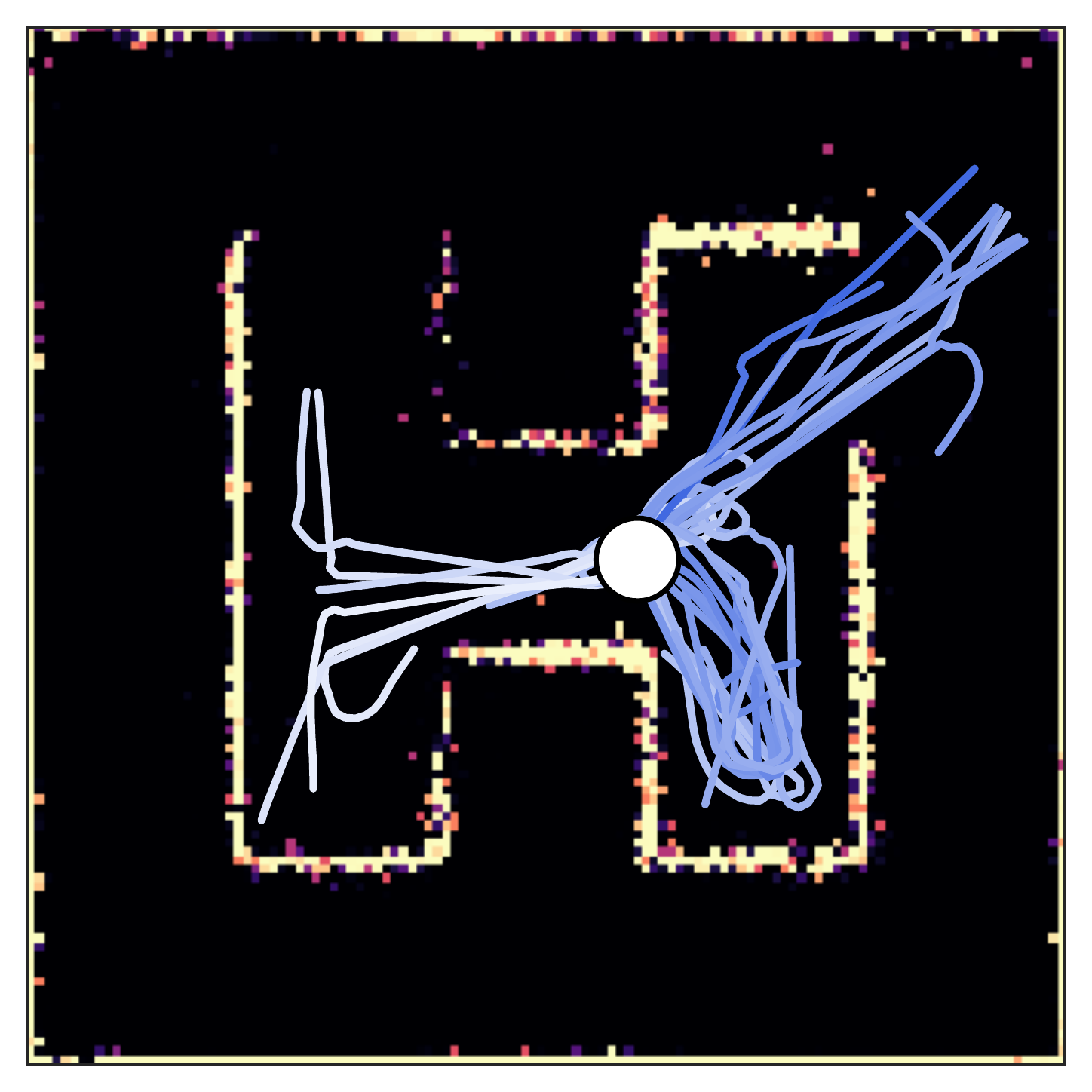}
		\caption{White means high MI score.}
		\label{subfig:slamcandidates1}
	\end{subfigure}
	\hfill
	\begin{subfigure}[t]{0.49\linewidth}
		\centering
		\small
		Map uncertainty \\
		\includegraphics[width=\linewidth]{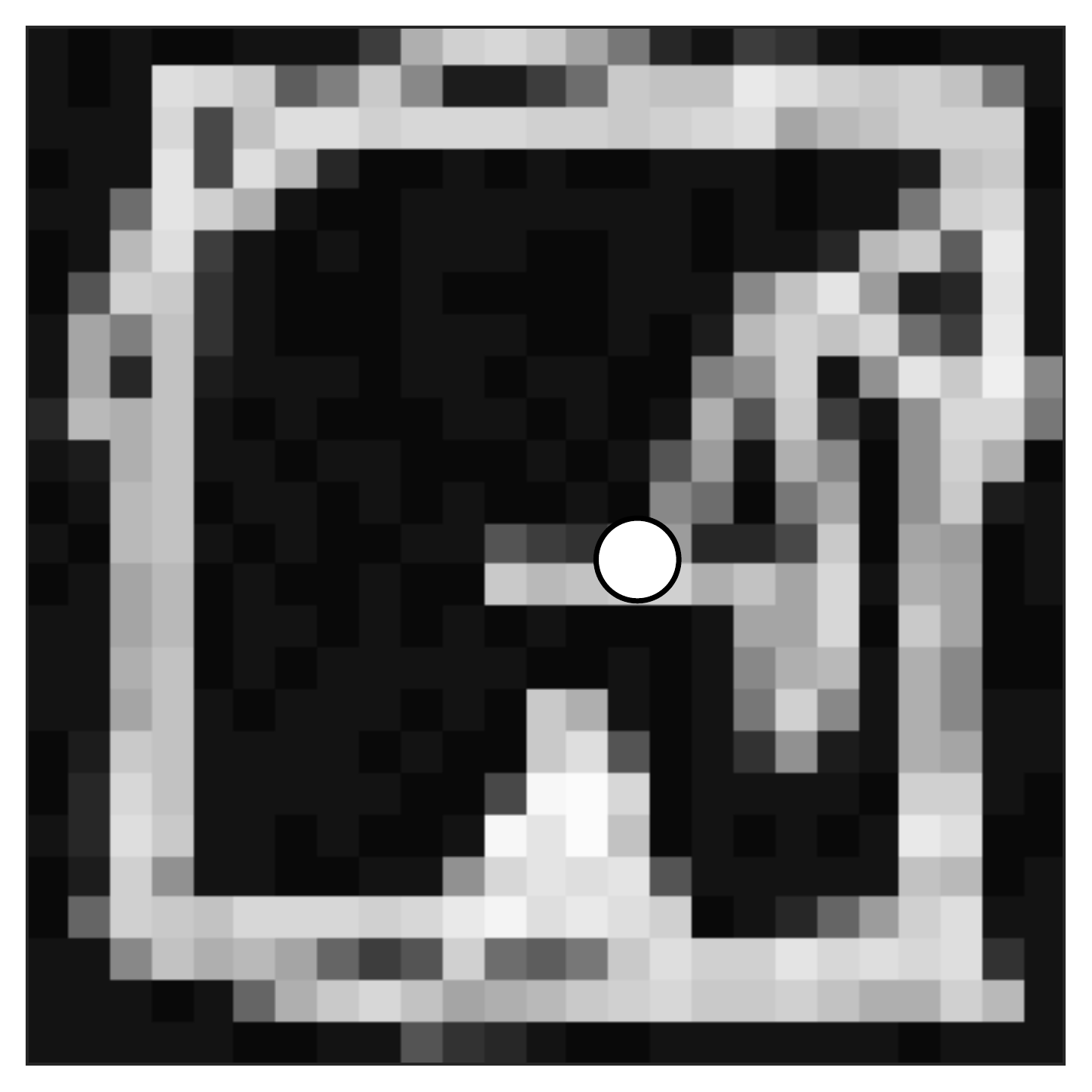}
		\caption{Black means high uncertainty.}
		\label{subfig:slamcandidates2}
	\end{subfigure}
	\caption{Generated candidates conform with the obstacles belief. Candidates leading into uncertain regions have high MI scores.}
	\label{fig:slamcandidates}
\end{minipage}%
\hfill
\begin{minipage}{.63\textwidth}
	\centering
	\includegraphics[width=\linewidth]{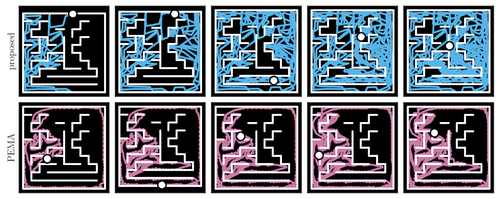}
	\caption{Qualitative exploration comparison: proposed method vs.\ LSTM baseline. The plot shows the parallel exploration progress of both agents over time.}
	\label{fig:explore}
\end{minipage}%
\end{figure*}

\section{\uppercase{EXPLORATION}}
\label{sec:exploration}
Next, we tackle the problem of exploration, expressed in the efficient mapping of the environment.
Fast inference of the map is a prerequisite for making informed decisions in spatial settings, as was demonstrated in the navigation task.  
Autonomous exploration amounts to the selection of control signals such that the data acquired makes inference progress fast.
The control signals are then executed in the environment and the process repeats.
\ifappendix{The complete procedure of our proposed exploration method is given in \Cref{alg:exploration} and the exact experimental details can be found in \Cref{apdx:exploration}.}
We will now discuss how we use DVBF-LM to define an exploration policy.

\subsection{Exploration via Active Learning}
We choose to follow an information-theoretic approach---we define optimal exploration to be that which leads to the largest change in information in the map variable, a metric commonly referred to as infogain \cite{infogain}.
The change in information is quantified by the mutual information (MI) between future observations $\bx\Ts$ predicted by DVBF-LM for a sequence of planned controls and the map $\Map$:
\begin{align*}
   &~ \expc[\bx\Ts \sim \p{\cdot}]{\kl{\p{\Map}{\bx\Ts}}{\p{\Map}}} \\
   =&~ \kl{\p{\bx\Ts, \Map}}{\p{\bx\Ts} \p{\Map}} \\
   =&~ I(\bx\Ts; \Map),  \numberthis \label{eq:infogain}
\end{align*}
omitting $\bu\Tsm$ for brevity.
Intuitively, the goal is to select those control signals which maximise the information gained through them in expectation.
Thus, we pose the following optimal-control problem:
\eq{
	\bu\Tsm^* = \arg \max_{\bu\Tsm} I(\bx\Ts; \Map | \bu\Tsm). \numberthis \label{eq:objective}
}
In light of this goal, the generative nature of DBVF-LM and the explicit modelling of a global latent map appear essential to the formulation of a principled exploration solution.  
To solve the posed problem, two issues need to be addressed: the intractability of computing mutual information from \Cref{eq:infogain} and conducting the optimisation in \Cref{eq:objective} (w.r.t.\ $\bu\Tsm$).

\subsection{Approximating Mutual Information}
\Cref{eq:infogain} depends on the highly nonlinear intractable joint $\p{\bx\Ts, \Map}{\bu\Tsm}$ and marginal $\p{\bx\Ts}{\bu\Tsm}$.
For brevity, we will omit the conditioning on $\bu\Tsm$ until the end of this section.
Approximation poses a challenge because of the double integration in the multi-dimensional spaces of $\bx\Ts$ and $\Map$.
We resort to combining MC sampling with applying a black-box entropy estimator, as done in~\cite{uddepeweg}.
This results in the following estimation:
\newcommand{\marginalX}{\mathbf{\bar{X}}}
\newcommand{\condX}{\mathbf{\tilde{X}}}
\eq{
	I(\Map; \bx\Ts) =&~ H[\bx\Ts] - H[\bx\Ts \mid \Map] \\
			      \approx&~\hat{H}(\marginalX)- \frac{1}{M} \sum_{m=1}^M \hat{H}(\condX^{(m)}). \numberthis \label{eq:ud}
}
$\hat{H}$ represents a black-box entropy estimator that works on sample sets.
$\marginalX$ is a set of samples from the marginal $\p{\bx\Ts}$ and each $\condX^{(m)}, m=1,\dots,M$ is a set of samples from a conditional $\p{\bx\Ts}{\Map^{(m)}}$ for $\Map^{(m)} \!\sim \q{\Map}$.
All samples are obtained through ancestral sampling from DVBF-LM, which is only possible since DVBF-LM is a generative model.

In practice, exploration is performed in parallel with the inferences of $\q{\bpose_{1:T}}$ and $\q{\Map}$, gradually adding new data points and increasing the overall data set size.
Note that we do not maximise MI once w.r.t.\ the prior $\p{\Map}$ for all future time steps, but we maximise it on-line multiple times for $T$ steps ahead w.r.t.\ the current variational posterior $\q{\Map} \approx \p{\Map}{\data}$.
This is well-grounded due to the validity of sequential Bayesian updates, \ie $\p{\Map}{\data, \bx\Ts} \propto \p{\bx\Ts}{\Map} \p{\Map}{\data}$.
In this work, a k-NN black-box entropy estimator is used for $\hat{H}$~\cite{ite}.
\subsection{Optimising Mutual Information}
Optimisation of \Cref{eq:objective} is performed in two stages.
First, a set of proposal controls $\mathcal{U}$ is generated by using the current belief of the map $\q{\Map} \approx \p{\Map}{\data}$.
Second, the best candidate control sequence $\bu\Tsm^* = \arg \max_{\bu\Tsm \in\, \mathcal{U}} I(\bx\Ts; \Map \mid \bu\Tsm, \data)$ is selected among the candidates and executed by the agent, following the scheme from the previous section.

Generating control sequences at random is very sample-inefficient, as the majority of sampled trajectories pass through obstacles in the environment.
Instead we follow a heuristic approach, exploiting the laser range nature of our sensors.
First we define an obstacle penalty $\lossobstacle(\bz)$ by building an occupancy map of the environment using $\q{\Map}{}$ and the DVBF-LM emission model.
We then draw $F$ random control sequences and form a set of candidates, minimising $\lossobstacle$ in expectation over the model:
\eq{
	&\bar\bu\Tsm^{(f)} = \arg\min_{\bu\Tsm} \expc[\bz\Ts \sim \p{\cdot}{\bu\Tsm, \data}]{\sum_{t=1}^T \lossobstacle(\bz_t)}, \numberthis \label{eq:obstacles}
}
with $f = 1, \dots, F$.
We then approximate \Cref{eq:objective} via
\eq{
	\bu\Tsm^* = \arg\max_{\bu^{(f)}\Tsm \in \mathcal{U}} I(\bx\Ts; \Map~|~\bu^{(f)}\Tsm).
}
\ifappendix{The obstacle penalty and overall generation procedure are described in more detail in \Cref{apdx:controlcandidate}.}
\Cref{subfig:slamcandidates1} illustrates the described steps.

\begin{figure}[t]
	\centering
	\includegraphics[width=\linewidth]{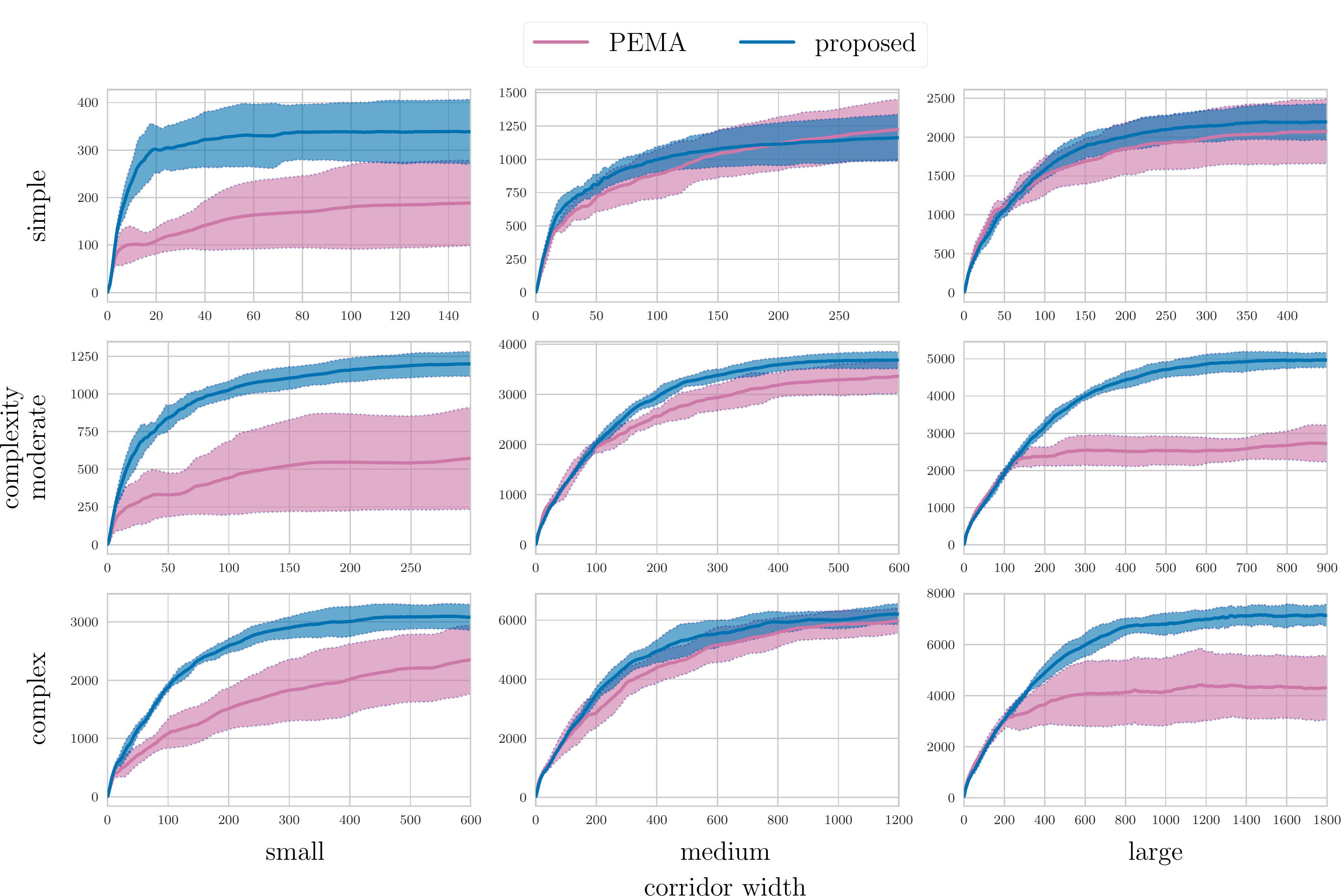}
	(a)
	\includegraphics[width=\linewidth]{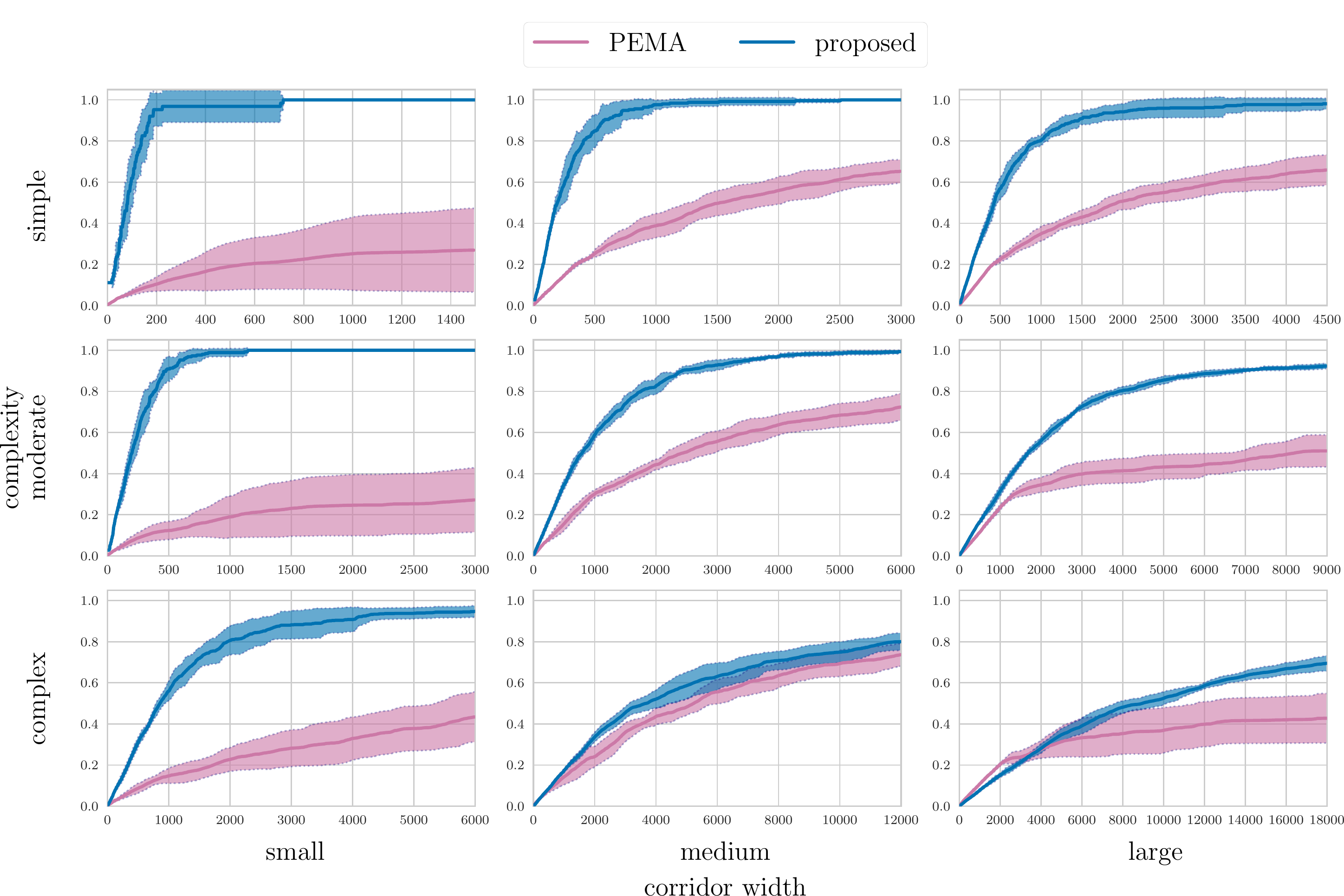}
	(b)
	\caption{
		Comparison of the proposed method (blue) against the LSTM baseline (violet): (a) infogain (b) exploration ratio.
		Higher values are better.
		The $x$ axis shows the number of interaction steps divided by 10.
		Shaded area corresponds to one standard deviation over seven runs.
	}
	\label{fig:quantresults}
\end{figure}

\subsection{Results}

The exploration experiments were carried out in our pybox2d simulator\ifappendix{(cf.\ \Cref{apdx:exploration} for details)}.
All model components are learned during exploration, with the exception of the transition model, which is handled in the same way as in the navigation experiments.
We evaluate explorative performance based on two metrics---information gain over the course of exploration and the fraction of ``tiles'' visited by the agent, the exploration ratio. 
For the latter, we divide the mazes into a fixed number of equally sized tiles\ifappendix{ (cf.\ \Cref{table:sizes})}.

As a baseline we consider a method that directly optimises the second metric, the exploration ratio. 
The baseline agent is represented by a deep deterministic LSTM network. 
Given a sequence of observations as its input, it puts out a control signal at each time step. 
We refer to it as the pose-entropy maximising agent (PEMA).
Since the exploration ratio objective is non-differentiable w.r.t. the LSTM parameters, we use Augmented Random Search \cite{ARS} to optimise it.
\ifappendix{For details on its objective and training, see \Cref{apdx:pema}.}

Qualitatively, we see that the proposed method rapidly traverses the maze. 
DVBF-LM exploration exhibits nearly uniform coverage, being driven to places that are visited least and are thus most uncertain (see \Cref{fig:slamcandidates}).
This holds even for the most complex mazes we considered, as we show in \Cref{fig:explore}.

The quantitative evaluation shows that our method consistently and significantly outperforms PEMA, even though PEMA is directly trained on the exploration ratio evaluation criterion. 
\Cref{fig:quantresults} summarises the comparison for the different metrics over time, aggregated over multiple runs in mazes with different complexity and corridor width.

\section{CONCLUSION AND FUTURE WORK}
We have introduced a deep variational Bayes filter that integrates a global latent variable of spatial form.
The novelty of our contribution lies in the flexibility that is inherited from neural networks and variational inference: contrary to most recent work in the area, our model still constitutes a generative model, which allows for a number of essential types of inference in spatial environments.
We validated the proposed method by applying it to the problems of SLAM, autonomous exploration and navigation.
Our model exhibits competitive localisation performance in comparison to an existing 2D LiDAR SLAM system, outperforms a strong baseline for exploration and can be used as a simulator for planning with virtually no loss in performance.
The results bear promise for real world application, which we will address in upcoming studies, along with comparisons to state-of-the-art visual SLAM methods.

\bibliographystyle{plainnat}
\bibliography{related}

\ifappendix{
    \appendices
    \crefalias{section}{appsec}
    \section{ENVIRONMENT DETAILS}
\subsection{Pybox2d Simulator for Mazes}
\label{sec:simulator}
\begin{figure}
	\label{fig:simulator}
	\begin{center}
	\includegraphics[height=3cm]{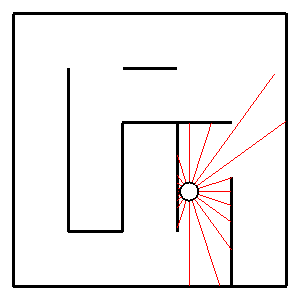}
	\end{center}
	\caption{
		Illustration of the maze environment. 
		The maze fills the unit square and is bounded by walls from all sides.
		The agent's pose $\bpose$ is its coordinates and rotation.
		The sensor readings are shown as lines emerging from the centre of the agent and go in all directions.
	}
\end{figure}
In this environment, an agent traverses a single square maze of side length $1$, which has been randomly generated.
The agent is modeled as a dynamic body that can move in the specified maze environment.
Its pose is specified by its coordinates in the maze's plane $\bcoord_t \in \mathbb{R}^2$ and its orientation $\Angle_t \in [-\pi, \pi]$, which we collect in $\bpose_t = (\bcoord_t, \Angle_t)$.
We assume the agent has a radius of $10^{-5}$.  
Additionally, its restitution parameter, which normally controls the bouncing off of objects after collisions with other bodies, is set to $0$. 
We define the agent's sensor to be a range finder with 20 line segments covering a full circle surrounding the agent ($\frac{2 \pi}{20}$ angular difference between neighboring beams); its response is the Euclidean distance to an object intersecting with the ray.
The length of each line segment is set to $0.53$, and that value is also returned when there is no obstacle in a given beam's reach. 
The agent's movement is restricted by collisions with the maze walls.
The agent is holonomic, as it can rotate freely with no obstruction, but can only move along the axis of its heading.
The control signals specify rotational velocity $\dot{\Angle}$ and a movement offset (directional derivative) $\dot{\bcoord}$. 
In the simulator, first the rotational velocity is applied, followed by the movement offset.

\begin{figure}
	\label{fig:mazes}
	\begin{center}
	\includegraphics[height=3cm]{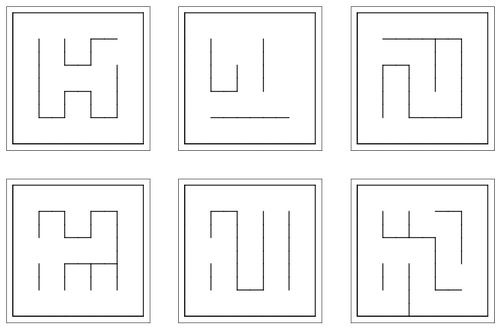}
	\end{center}
	\caption{The six random mazes used for evaluation.}
\end{figure}

\subsection{VizDoom Environment}
\label{sec:vizdoomenv}
This environment is implemented using the popular VizDoom simulator, in which the agent is identifiable with a player traversing a level of the game Doom.
Doom level files (WADs) are generated for every square maze pattern (the same ones used in pybox2d), such that the maze walls are specified as vertical obstacles with no thickness.
We use the default VizDoom textures and we do not add any characteristic patterns to the walls.
The agent's pose is specified by its coordinates $\bcoord_t \in \RR^2$ on the given maze level and its orientation $\alpha_t \in [-\pi, \pi]$, which we collect in $\bpose_t = (\bcoord_t, \alpha_t)$.
For convenience, we normalise the level coordinates to $[0, 1]^2$.
In contrast to the pybox2d environment, in this case the observations collected by the agent are images depicting the player's view at every moment of interaction.
The screen resolution is set to $320 \times 240$ pixels with $24$-bit colors.
All user interface elements, the player's weapon, any screen messages or effects are disabled, so that only the surrounding maze structures are visible.
The collected RGB images are converted to grayscale and scaled down to $32 \times 24$ pixels for the sake of computational efficiency.
Pixel intensities are normalised to the $[0, 1]$ range.
We do not use the available depth buffer.
The agent can rotate freely, but it's movement is restricted by collisions with the maze walls.
The available control signals specify rotational velocity $\dot{\alpha}$ and movement offset $\dot{\bcoord}$ in the direction of the agent's heading.
We implemented these by executing TURN\_LEFT, TURN\_RIGHT and MOVE\_FORWARD actions in VizDoom for a given number of game ticks.

\begin{figure}
	\begin{center}
	\begin{subfigure}[t]{\linewidth}
		\centering
		\includegraphics[height=3.0cm]{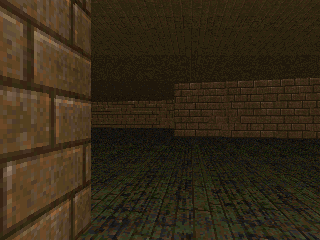}
		\caption{}
	\end{subfigure}
	\begin{subfigure}[t]{\linewidth}
		\centering
		\includegraphics[height=3.0cm]{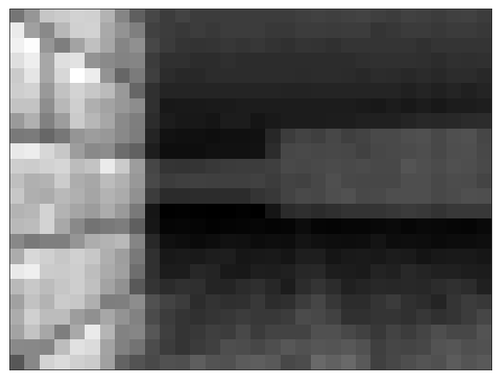}
		\caption{}
	\end{subfigure}
	\end{center}
	\caption{
		Examples for observations obtained in the VizDoom environment. 
		(a) Original $320 \times 240$ RGB image from the simulator.
		(b) Processed $32 \times 24$ grayscale version of the same image. 
	}
	\label{fig:qualitative}
\end{figure}

\section{DETAILS OF THE DVBF-LM SLAM MODELS}
\label{sec:details}

\begin{table}
	\caption{Network Architecture Details}
	\vspace{1eM}
	\centering
	\begin{tabular}{llll}
		\toprule
		model                     & \# layers   & \# units   & activation \\
		\midrule
		emission, $\genpars_E$   & 4                   & 256       & softsign \\
		transition, $\genpars_T$ & 6                   & 256       & ReLU \\
		\bottomrule
	\end{tabular}
	\label{table:architectures}
\end{table}

\subsection{Emission Model}
\begin{itemize}
	\item Input: $\bmap_t$
	\item Output: $\bobs_t$
	\item $\bmu_E$ is realised via a feed-forward neural network with parameters $\genpars_E$ (cf.\ Table~\ref{table:architectures} for hyperparameters)
	\item The emission mean $\bmu_{\bobs_t}$ is formed by applying a rotational shift to the output of $\bmu_E(\bmap_t)$ w.r.t.\ the current estimate of the agent's heading angle $\phi_t = \bpose_t^{(1)}$; this is done to ensure the generalisation of charts $\bmap_t$ for all different orientations of the agent
	\item The emission model's standard deviation is globally set to $0.1$: \newline

		\centering{$\bobs_t = \mcN(\bmu_{\bobs_t}, 0.1 \times \mathbf{I})$}
\end{itemize}

\subsection{Transition Model}
\begin{itemize}
	\item Input: $\bpose_t, \bcontrol_t, \bmap_t$
	\item Output: $\bpose_{t+1}$
	\item $\bmu_T$ is realised via a feed-forward neural network with parameters $\genpars_T$ (cf.\ Table~\ref{table:architectures} for hyperparameters)
	\item To avoid potential discontinuities in angular space, we transform $\phi_t = \bpose_t^{(1)}$ (the agent's heading at the current time step) to $\left [ \cos \phi_t, \sin \phi_t \right ]^T$ before applying the transition model
	\item Instead of feeding $\bmap_t$ into $\bmu_T$ directly, we choose to feed in $\bmu_E(\bmap_T)$ (the mean of the emission model, a deterministic transformation of $\bmap_t$) --- this allows us to pre-train the transition network using ground truth observations.
	\item In our experiments, we assume no noise in the transition model, i.e. $\sigma_T^2 \rightarrow 0$.
\end{itemize}

\subsection{Memory}
\begin{itemize}
	\item $\Map \in \mathbb{R}^{w \times h \times D_m}$, $w = h = 32$, $D_m = 10$
	\item Initialisation of the parameters of $\q{\Map}$ (inference):

		\centering{$\mathbf{\phi}_{\Map} = \{ \bmu_{\Map_{ij}}, \bsigma^2_{\Map_{ij}} \}_{i \in [w], j \in [h]}, \bmu_{\Map_{ij}} = \mathbf{0}, \bsigma^2_{\Map_{ij}} = \mathbf{1}$}
\end{itemize}

\subsection{Emission Model (VizDoom)}
\label{sec:vizdoomemitdetails}
\begin{itemize}
	\item Input: $\bmap_t$, of shape $32\times24$
	\item Output: $\bobs_t$, of shape $32\times24$
	\item The input $\bmap_t$ is rotated with respect to the agent's current heading angle and cropped according to the known field-of-view of the camera, which is $\ang{90}$.
	\item $\bmu_E$ is realised by a set of $16$ feed-forward neural networks. Each network translates one row of $\bmap_t$ into one row of the reconstructed image. For the top $16$ rows we used networks $1-16$ and for the bottom sixteen rows we used the same networks in reversed order ($16-1$).
	\item The number of layers and hidden units are the same as for the pybox2d experiments. We used rectified linear units.
	\item The emission model's standard deviation is globally set to $0.1$.
\end{itemize}

\subsection{Transition Model (VizDoom)}
\begin{itemize}
	\item We used the same configuration as for the pybox2d experiments.
\end{itemize}

\subsection{Memory (VizDoom)}
\begin{itemize}
	\item $\Map \in \mathbb{R}^{w \times h \times k \times l}$, $w = h = 25$, $k = 32$, $l = 24$
	\item Same initialisation as in the pybox2d setting.
\end{itemize}

\subsection{Approximate Asynchronous Particle Representation}
\begin{itemize}
	\item Number of particles: 50
	\item Update frequency: particles are updated every 50 mini-batches
\end{itemize}

\subsection{Optimization}

\begin{itemize}
	\item Optimizer: Adam, $\beta_1 = 0.9$, $\epsilon = 10^{-8}$
	\item Learning rate: $10^{-4}$
	\item Batch-size: $128$
\end{itemize}

\subsection{Particle Filtering in the Offline Setting}

The particle filter used in the offline setting consists of the variational approximation $\q{\bpose_{t}}$. There are two notable differences:

\begin{itemize}
	\item The individual chunks of the trajectory are processed sequentially.
	\item The proposal distribution $\hat q$ is a categorical distribution over the set of particles (as opposed to a moment-matched Normal distribution).
\end{itemize}

The particle filtering routine has four hyper-parameters:

\begin{itemize}
	\item The length of each chunk: $5$
	\item The amount of noise added to the particles at the beginning of each chunk: $0.001$
	\item The number of particles: $1000$
	\item The standard deviation of the emission distribution $\bsigma_E = 0.01$ 
\end{itemize}

These were tuned by random search on the same sequence that was used for early-stopping when training the transition model parameters.

    \section{NAVIGATION EXPERIMENTS}
\subsection{Safety Term for Navigation}
\label{sub:safety}
The safety term $\mathcal{L}_{\mathrm{safe}}=\sum_{i}s(l_i)$ depends on the parametric sigmoidal function $s(\cdot)$.
In our experiments we use the following parameterisation for $s(\cdot)$:
\eq{
	s(x) = \dfrac{\mu}{1 + \exp{((x - \delta) * \sigma)}}.
}
This term is designed to rapidly increase as soon as the lidar reading $x$ drops below a threshold $\delta$.
The parameters $\mu$, $\delta$ and $\sigma$ are configured as:
\begin{itemize}
\item $\mu = 10^4$
\item $\delta = 0.03$
\item $\sigma = 10^2$
\end{itemize}

    \section{EXPLORATION EXPERIMENTS}\label{apdx:exploration}

\ifnotappendix{\newcommand{\marginalX}{\mathbf{\bar{X}}}}
\ifnotappendix{\newcommand{\condX}{\mathbf{\tilde{X}}}}
\begin{figure}[t]
	\centering
	\includegraphics[width=0.7\linewidth]{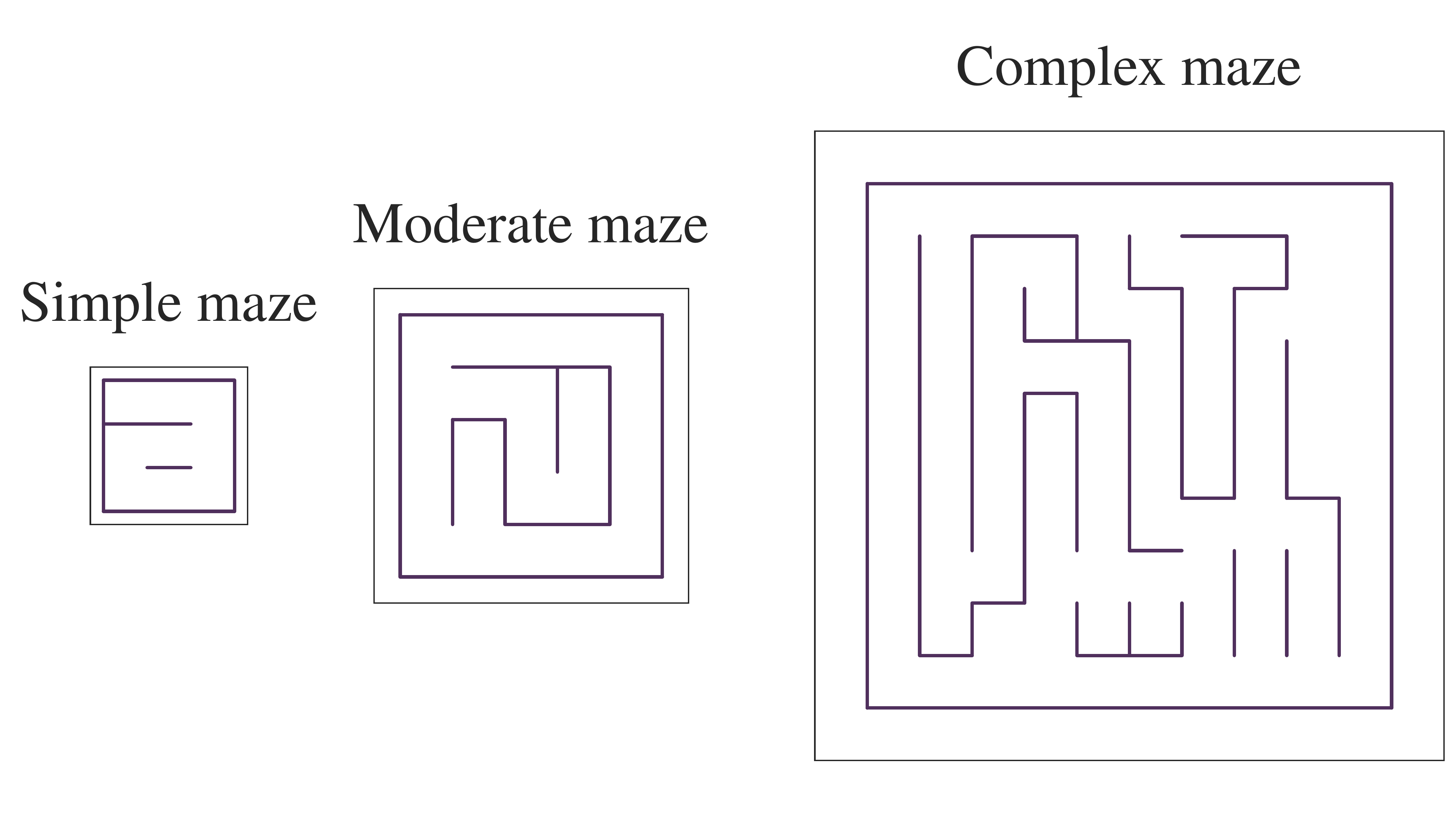}
	\caption{Example mazes for the three considered maze complexities for exploration.}\label{fig:mazecomplexity}
\end{figure}

\begin{table*}[h]
	\small
	\caption[Details about the explored mazes.]{Exploration scale matrix for the conducted experiments, describing the differences in map variable dimensions, physical maze sizes, allowed exploration steps and number of tiles used to evaluate exploration ratios for the different exploration scenarios.} \label{table:sizes}
	\vspace{1eM}
	\centering
	\begin{tabular}{llcccc}
		\toprule
		complexity & corridor width & map dimensions & maze size & steps & evaluation tiles \\
			   & & ($w \times h \times D_{\bm}$) & & \\
		\midrule
		simple & small & $7 \times 7 \times 20$ & $0.25 \times 0.25$ & $1500$ & $3 \times 3$ \\
		\midrule
		simple & medium & $12 \times 12 \times 20$ & $0.5 \times 0.5$ & $3000$ & $6 \times 6$ \\
		\midrule
		simple & large & $18 \times 18 \times 20$ & $0.75 \times 0.75$ & $4500$ & $9 \times 9$ \\
		\midrule

		moderate & small & $12 \times 12 \times 20$ & $0.5 \times 0.5$ & $3000$ & $6 \times 6$ \\
		\midrule
		moderate & medium & $25 \times 25 \times 20$ & $1.0 \times 1.0$ & $6000$ & $12 \times 12$ \\
		\midrule
		moderate & large & $37 \times 37 \times 20$ & $1.5 \times 1.5$ & $9000$ & $18 \times 18$\\
		\midrule

		complex & small & $25 \times 25 \times 20$ & $1.0 \times 1.0$ & $6000$ & $12 \times 12$ \\
		\midrule
		complex & medium & $50 \times 50 \times 20$ & $2.0 \times 2.0$ & $12000$ & $18 \times 18$ \\
		\midrule
		complex & large & $75 \times 75 \times 20$ & $3.0 \times 3.0$ & $18000$ & $24 \times 24$ \\
		\bottomrule
	\end{tabular}
\end{table*}

The exploration experiments are conducted with the same pybox2d simulator used for the laser range finder SLAM experiments (cf. \cref{sec:simulator}).
Additionally, we consider three different maze complexity levels, single examples of which are depicted in \cref{fig:mazecomplexity}.
For each complexity level, we instatiate multiple different random mazes varying the overall maze size, and thus the corridor widths.
As a reference, the side length of the mazes with moderate complexity and average corridor width is set to $1.0$.
The proposed active SLAM method is then performed in each generated maze.
The number of allowed time steps varies with the maze scale.
The exact details on the considered mazes are given in \cref{table:sizes}.

For the evaluation, the reported infogain over time is the KL divergence between the initial memory prior $p(\Map)$ and the approximation $q(\Map)$ at the time of measurement:
\eq{
	\kl{q(\Map)}{p(\Map)}.
}
The reported exploration ratio is the ratio of visited tiles at the time of measurement, where we vary the overall number of tiles w.r.t. the scale of each individual maze (cf. \cref{table:sizes}).

\subsection{Active SLAM Model Parameters}
\begin{itemize}
	\itemsep0em
	\item \emph{Emission}: $\genparam_E$ is learned from data during exploration and represents the weights of a feed-forward neural network with $4$ layers, $256$ units, softsign activation. The emission standard deviation is $\sigma_E = 0.1$.
	\item \emph{Transition}: $\genparam_T$ is assumed to be known for the conducted exploration (specified by the user) with standard deviation $\sigma_T = 0.0004$ per time step.
	\item \emph{Map}: The depth dimension of the map variable is fixed: $D_{\bm} = 20$. The width $w$ and $h$ vary based on the size of the explored maze (cf.\ \Cref{table:sizes}). The inference $\q[\infparam]{\map}$ is initialised as a standard normal distribution (like the prior).
	\item \emph{Particle Filtering}: The employed particle filter uses $20$ particles to represent the posterior over poses.
	\item \emph{Optimisation}: When the model is trained, the data is split up into chunks of $5$ steps. A batch size of $128$ chunks is used. The chosen optimiser is Adam, with $\beta_1 = 0.9, \epsilon = 10^{-8}$ and a learning rate of $0.001$.
\end{itemize}

\begin{algorithm*}
	\caption{Exploration of global parameters in deep state-space models}\label{alg:exploration}
\begin{algorithmic}[1]
	\Require{
		\\\p[\text{env}]{\bx\Ts}{\bu\Ts, \data} --- the empirical distribution of the environment
		\\$\genpars_0, \varpars_0$ --- initial parameters
		\\$N_{\text{train}}$ --- overall number of training steps
	}
	\Statex
	\Function{Explore}{$\p[\text{env}], \genpars_0, \varpars_0, N_{\text{train}}$}
	\Let{$\genpars, \varpars$}{$\genpars_0, \varpars_0$} 
	\Comment{initialise generative and inference parameters ($\genpars_T, \genpars_E$, etc.)}
	\Let{$\data$}{$\emptyset$}
	\Comment{start with an empty data set}
	\Loop
	\Let{$\mathcal{U}$}{\Call{GenerateCandidates}{$\data, \genpars, \varpars$}}
	\Comment{set of possible control candidates}
	\Let{$\bu\Tsm^*$}{$\arg \max_{\bu\Tsm \in \mathcal{U}} I(\globalvar; \bx\Ts \mid \bu\Tsm, \data)$}
	\Comment{optimal controls}
	\State $\bx\Ts \sim \p[\text{env}]{\,\cdot}{\bu\Tsm^*, \data}$
	\Comment{execute the optimal exploration controls}
	\Let{$\data$}{$\data \cup \{ \left( \bx\Ts, \bu\Tsm^* \right) \}$}
	\Comment{extend the data set}

	\For{$i \gets 1 \textrm{ to } N_{\text{train}}$}
	\Comment{train on the current data set}
	\Let{$\Delta \genpars, \Delta \varpars$}{$\nabla_{\genpars, \varpars} \loss[\text{ELBO}]{\data, \genpars, \varpars}$}
	\Let{$\genpars, \varpars$}{$\genpars + \eta \Delta \genpars, \varpars + \eta \Delta \varpars$}
	\EndFor

	\EndLoop
	\EndFunction
\end{algorithmic}
\end{algorithm*}

\subsection{Mutual Information Exploration Parameters}
\begin{itemize}
	\itemsep0em
	\item The exploration horizon is $T = 50$ steps long. This means that new controls are selected every 50 steps by reestimating MI.
	\item The number of considered candidate trajectories $\bu\Tsm$ is $F = 40$.
	\item For the estimation of mutual information based on the estimator $\hat{H}(\marginalX)- \frac{1}{M} \sum_{m=1}^M \hat{H}(\condX^{(m)})$, $L = 30$ samples are used for the estimation of the marginal predictive entropy term, $M = 40$ samples are used for the MC estimation of the outer expectation in the conditional entropy term, and $K = 100$ samples are used for the estimation of the entropies inside the conditional entropy term.
	\item Exploration (selection and execution of $T$ new control inputs) is performed after every $500$ successfully executed SLAM training steps.
\end{itemize}

    \section{DETAILS ON CONTROL CANDIDATE GENERATION}\label{apdx:controlcandidate}

\begin{algorithm*}[t]
	\caption{Generating candidate control trajectories}\label{alg:candidates}
\begin{algorithmic}[1]
	\Require{
		\\$\data = (\bx_{1:N}, \bu_{1:N-1})$ --- previously collected data
		\\$\genpars, \varpars$ --- DVBF-LM parameters
	}
	\Statex
	\Function{GenerateCandidates}{$\data, \genpars, \varpars$}
	\Let{$\hat{\Map}$}{mean (and mode) of $\q[\varpars]{\Map}$}
	\Let{$\hat{\mathbf{z}}_{1:N}$}{mean of all particles from $\q[\genpars]{\bz_{1:N}}{\bx_{1:N}, \bu_{1:N-1}, \hat{\Map}}$}
	\Let{$\hat{\mathbf{x}}_{1:N}$}{mean (and mode) of $\p[\genpars]{\bx_{1:N}}{\hat{\mathbf{z}}_{1:N}, \hat{\Map}}$}
	\Let{\p[\text{obstacle}]{\bz}{\hat{\mathbf{z}}_{1:N}, \hat{\mathbf{x}}_{1:N}}}{estimated density of obstacle based on $\hat{\mathbf{z}}_{1:N}$ and $\hat{\mathbf{x}}_{1:N}$}
	\Let{$\loss[\text{obstacle}]{\bz}$}{bilinear interpolation of $\p[\text{obstacle}]{\bz}{\hat{\mathbf{z}}_{1:N}, \hat{\mathbf{x}}_{1:N}}$}
	\Let{$\mathcal{U}$}{$\emptyset$}
	\For{$f \gets 1 \textrm{ to } F$}
	\Let{$\bu\Tsm^{(f)}$}{$\arg\min_{\bu\Tsm} \expc[\bz\Ts \sim \p{\,\cdot}{\bu\Tsm, \data}]{\sum_{t=1}^T \loss[obstacle]{\bz_t}}$}
	\Let{$\mathcal{U}$}{$\mathcal{U} \cup \{\bu\Tsm^{(f)}\}$}
	\EndFor
	\State \Return{$\mathcal{U}$}
	\EndFunction
\end{algorithmic}
\end{algorithm*}
The generation of control sequence candidates happens every time before the mutual information needs to be evaluated in order to explore the environment.
A wall penalty built by means of the current map belief $\q{\Map}$ is used for the optimisation of:
\eq{
	&\bar\bu\Tsm^{(f)} = \arg\min_{\bu\Tsm} \expc[\bz\Ts \sim \p{\cdot}{\bu\Tsm, \data}]{\sum_{t=1}^T \loss[obstacle]{\bz_t}}, \numberthis \label{eq:obstacles}
}
Since the observations are laser range readings, the predicted observation values at different regions of the map are used to form a probability measure for the presence of obstacles.
Let $\bx_{1:N}, \bu_{1:N-1}$ denote all previously collected laser readings and corresponding controls.
In other words, $\data = (\bx_{1:N}, \bu_{1:N-1})$, where $\data$ denotes all previously collected data.
Let $\hat{\mathbf{z}}_{1:N}$ denote the mean of all particle sets used to represent $\q{\bz_{1:N}}{\bx_{1:N}, \bu_{1:N-1}, \Map}$, the approximate posterior over all previous poses in the DVBF-LM model.
Let $\hat{\mathbf{x}}_{1:N}$ denote the mean (and simultaneously mode, because of the Gaussian emission assumption) of $\p{\bx_{1:N}}{\hat{\mathbf{z}}_{1:N}, \Map}$.
Hence, $\hat{\mathbf{x}}_{1:N}$ correspond to the predicted laser readings, based on the current map belief, at the mean belief of all previously visited locations expressed by $\hat{\mathbf{z}}_{1:N}$.
We then interpret the probability of the presence of an obstacle at position $\bz$ as:
\begin{align*}
	&\log \p[\text{obstacle}]{\bz}{\hat{\mathbf{z}}_{1:N}, \hat{\mathbf{x}}_{1:N}} \\
	&\begin{aligned}
		\propto \frac{1}{N K_{\text{laser}}} \sum_{n=1}^{N} \sum_{k=1}^{K_{\text{laser}}} \sum_{s_{ij} \in \mathcal{S}} \mathbb{I}(f_{\text{laser}}(\hat{\mathbf{z}}_n, \hat{\mathbf{x}}_n^{(k)}) &\in s_{ij}) \\
		&\cdot \mathbb{I}(\bz \in s_{ij}).
	\end{aligned}
\end{align*}
In the above, $s_{ij} \in \mathcal{S}$ denotes one of $W \times H$ discrete square cells spanning the 2D area of the whole maze in a grid.
Correspondingly, $\mathbb{I}$ is the indicator function which returns $1$ if a position $\bz$ is inside a cell $s_{ij}$ and $0$ otherwise.
Finally, $f_{\text{laser}}(\bz, \bx^{(k)})$ simply evaluates the 2D position of an obstacle measured by the $k$-th laser of a reading obtained at $\bz$.
In other words, the probability density for the presence of an obstacle is expressed as a histogram of projected points based on the laser predictions for all previously collected data.
Because the defined probability density is a step function, bilinear interpolation is applied over $\log \p[\text{obstacle}]{\bz}{\hat{\mathbf{z}}_{1:N}, \hat{\mathbf{x}}_{1:N}}$ to allow for direct gradient based optimisation.
$\loss[\text{obstacle}]{\bz}$ denotes the function of the resulting loss landscape.
A depiction of an example evaluation of the loss surface from an actual experiment can be found in \Cref{fig:wallloss}.
The loss is then minimised $F$ times, generating different candidate control sequences every time due to random initialisation:
\begin{align*}
	&\bu\Tsm^{(f)} \\
	&= \arg\min_{\bu\Tsm} \expc[\bz\Ts \sim \p{\cdot}{\bu\Tsm, \data}]{\sum_{t=1}^T \loss[\text{obstacle}]{\bz_t}}
\end{align*}
\begin{figure}[t]
	\begin{subfigure}[t]{0.32\linewidth}
		\centering
		Probability density of walls
		\includegraphics[width=\linewidth]{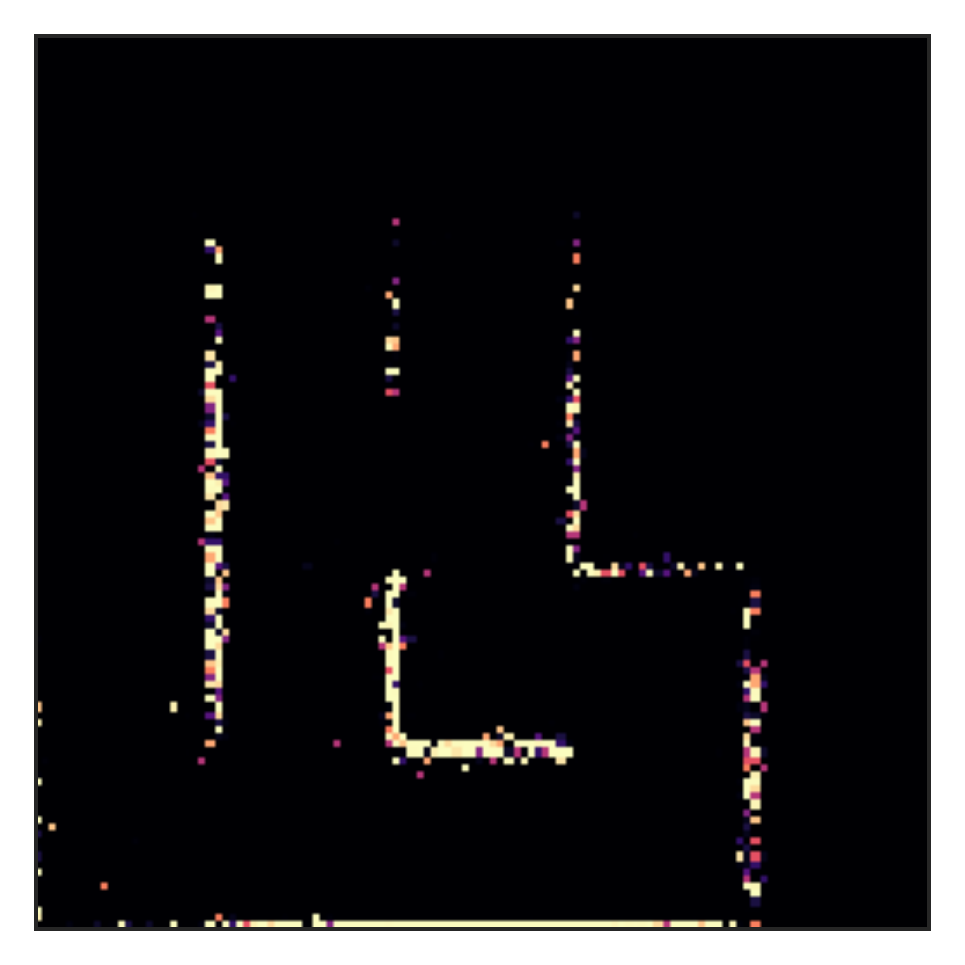}
	\end{subfigure}
	\hfill
	\begin{subfigure}[t]{0.32\linewidth}
		\centering
		Interpolated wall loss
		\includegraphics[width=\linewidth]{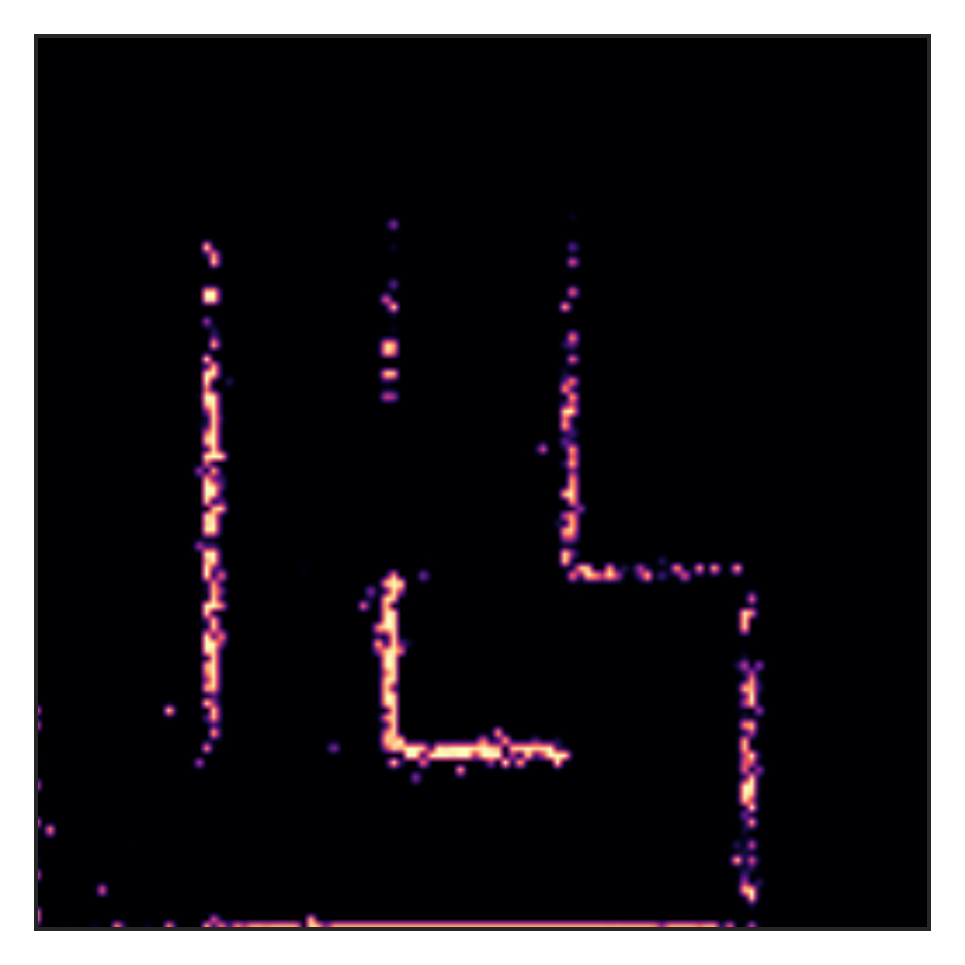}
	\end{subfigure}
	\hfill
	\begin{subfigure}[t]{0.32\linewidth}
		\centering
		Corresponding traversal
		\includegraphics[width=\linewidth]{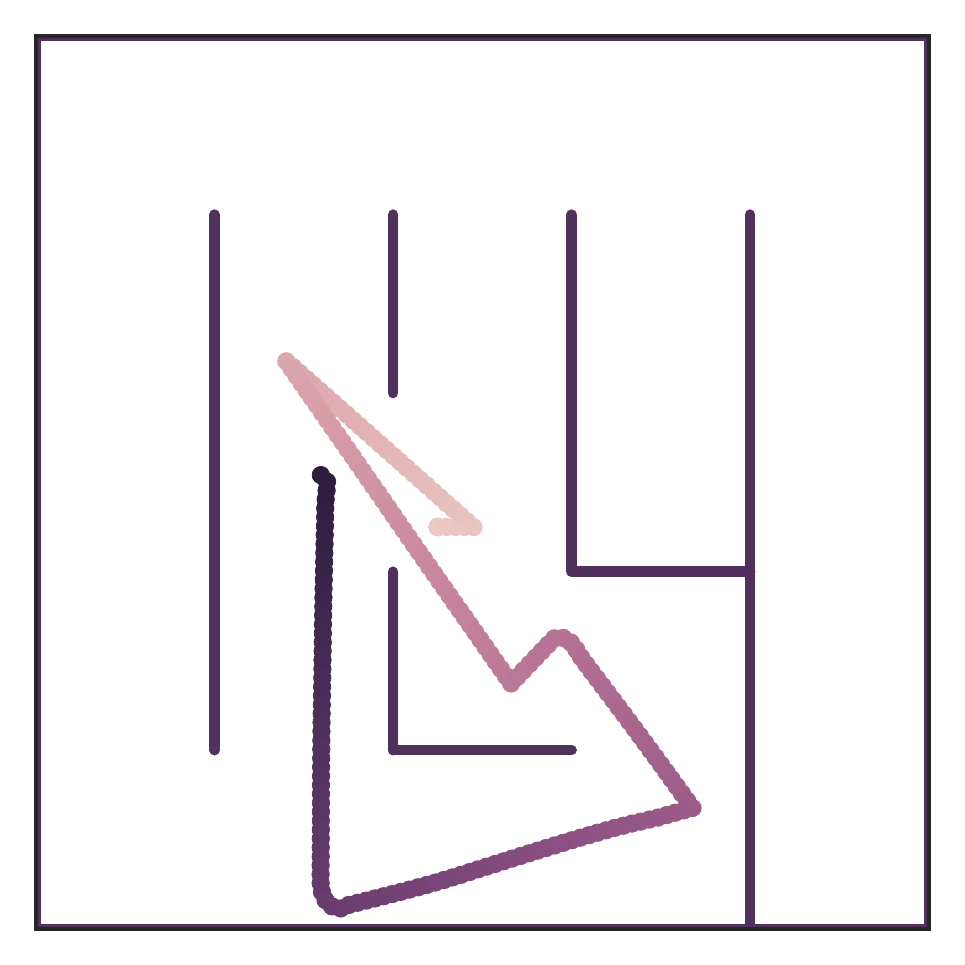}
	\end{subfigure}
	\caption[Candidate controls generation loss.]{An illustration of the resulting loss landscape for the generation of candidate control sequences for which MI should be estimated.
	In the above depiction, the previously collected data $\data$ does not cover the whole maze, hence why the belief of obstacles only partially covers the ground truths.}\label{fig:wallloss}
\end{figure}
where $\p{\bz\Ts}{\bu\Tsm, \data}$ is the predictive DVBF-LM model over agent positions given controls $\bu\Tsm$.
Sampling from it is done by consecutive application of the transition with parameters $\genpars_T$, utilising only the current map belief $\q{\Map}$ without observing the environment for the controls $\bu\Tsm$ (i.e.\ no actual rollouts are necessary).
The overall generation procedure is described in \Cref{alg:candidates}.

\subsection{Candidate Control Generation Parameters}
\begin{itemize}
	\itemsep0em
	\item \emph{Optimisation}: \Cref{eq:obstacles} is optimized via gradient descent for $500$ update steps. All $40$ candidate controls sequences, each of $100$ steps, are optimised in parallel. The chosen optimiser is Adam and the learning rate is set to $0.001$.
	\item \emph{Regularisation}: To promote the generation of trajectories that do not circle in place, an L2 regularisation term is applied to the resulting pose angles $\varphi_t$. A hyperparameter search revealed $2.0$ to be the optimal regularisation coefficient.
\end{itemize}

    \section{POSE-ENTROPY-MAXIMISING AGENT}\label{apdx:pema}

\begin{algorithm*}
	\caption{Execution of a PEMA rollout}\label{alg:pemaexec}
\begin{algorithmic}[1]
	\Require{
		\\ $\genparam_{\text{pema}}$~--- weights of the PEMA LSTM
		\\ $p_{\text{env}}(\bz,\bx \mid \bu, \data)$~--- empirical distribution of an environment providing ground truth poses $\bz$
		\\ $T$~--- number of execution steps
		}
	\Statex
	\Function{ExecuteRollout}{$\genparam_{\text{pema}}, p_{\text{env}}, T$}
	\Let{$\bu$}{$\mathbf{0}$}
	\Let{$\data$}{$\emptyset$}
	\Let{$\mathcal{Z}$}{$\emptyset$}
	\Comment{list of collected true poses}
	\For{$t \gets 1 \textrm{ to } T$}
		\State $\bz, \bx \sim \p[\text{env}]{\,\cdot}{\bu, \data}$
		\Comment{execute control for one step}
		\Let{$\bu$}{$\text{LSTMCell}_{\genparam_{\text{pema}}}(\bx)$}
		\Comment{apply the LSTM cell to obtain next control}
		\Let{$\mathcal{Z}$}{$\mathcal{Z} \cup \left \{ \bz \right \}$}
		\Let{$\data$}{$\data \cup \left \{ (\bx, \bu, \bz) \right \}$}
	\EndFor
	\State \Return{$\mathcal{Z}$}
	\Comment{return the list of collected true poses}
	\EndFunction
\end{algorithmic}
\end{algorithm*}

\begin{algorithm*}
	\caption{PEMA baseline training procedure}\label{alg:pemaalg}
\begin{algorithmic}[1]
	\Require{
		\\ $\left \{p_{\text{env}}^{(m)}(\bz,\bx \mid \bu, \data) \mid m \in [M] \right \}$ --- empirical distributions of the $M$ training environments with access to true poses $\bz$
		\\ $K_{\text{pert}}$~--- number of weight perturbations
		\\ $T$~--- number of execution steps
		\\ $\genparam_{\text{pema}}^0$~--- initial LSTM weights
		\\ $\eta$~--- update step length
		}
	\Statex
	\Function{TrainBaseline}{$\left \{p_{\text{env}}^{(m)} \mid m \in [M] \right \}, K_{\text{pert}}, T, \genparam_{\text{pema}}^0, \eta$}
	\Let{$\genparam_{\text{pema}}$}{$\genparam_{\text{pema}}^0$}
	\Comment{initialise LSTM weights}
	\While{not converged}
		\Let{$\Theta_{\text{back}}$}{$\emptyset$}
		\Comment{set of backward perturbed weights}
		\Let{$\Theta_{\text{forth}}$}{$\emptyset$}
		\Comment{set of forward perturbed weights}
		\Let{$\mathcal{S}_{\text{pert}}$}{$\emptyset$}
		\Comment{set of perturbation values}
		\Let{$\mathcal{S}_{\text{score}}$}{$\emptyset$}
		\Comment{set of scores for each perturbation}
		\For{$k \gets 1 \textrm{ to } K_{\text{pert}}$}
			\State $\boldsymbol{\epsilon} \sim \gauss{\mathbf{0}}{\sigma_{\text{pert}}^2\mathbf{I}}$
			\Comment{perturb weights for ARS}
			\Let{$\genparam_{\text{back}}$}{$\genparam_{\text{pema}} - \mathbf{\epsilon}$}
			\Let{$\genparam_{\text{forth}}$}{$\genparam_{\text{pema}} + \mathbf{\epsilon}$}

			\Let{$s_{\text{back}}$}{$0$}
			\Comment{score for the backward perturbation}
			\Let{$s_{\text{forth}}$}{$0$}
			\Comment{score for the forward perturbation}
			\For{$m \gets 1 \textrm{ to } M$}
				\Comment{obtain scores in all training environments}
				\Let{$\bz\Ts^{\text{back}}$}{\Call{ExecuteRollout}{$\genparam_{\text{back}}, p_{\text{env}}^{(m)}, T$}}
				\Comment{cf.\ \Cref{alg:pemaexec} above}
				\Let{$\bz\Ts^{\text{forth}}$}{\Call{ExecuteRollout}{$\genparam_{\text{forth}}, p_{\text{env}}^{(m)}, T$}}
				\Let{$s_{\text{back}}$}{$s_{\text{back}} - \loss[\text{pema}]{\bz\Ts^{\text{back}}}$}
				\Comment{objective from \Cref{eq:pemaloss}}
				\Let{$s_{\text{forth}}$}{$s_{\text{forth}} - \loss[\text{pema}]{\bz\Ts^{\text{forth}}}$}
			\EndFor

			\Let{$\Theta_{\text{back}}$}{$\Theta_{\text{back}} \cup \left \{ \genparam_{\text{back}} \right \}$}
			\Let{$\Theta_{\text{forth}}$}{$\Theta_{\text{forth}} \cup \left \{ \genparam_{\text{forth}} \right \}$}
			\Let{$\mathcal{S}_{\text{pert}}$}{$\mathcal{S}_{\text{pert}} \cup \left \{ \mathbf{\epsilon} \right \}$}
			\Let{$\mathcal{S}_{\text{score}}$}{$\mathcal{S}_{\text{score}} \cup \left \{ (s_{\text{back}}, s_{\text{forth}}) \right \}$}
		\EndFor

		\Let{$\Delta \genparam_{\text{pema}}$}{\Call{ARSStep}{$\Theta_{\text{back}}, \Theta_{\text{forth}}, \mathcal{S}_{\text{pert}}, \mathcal{S}_{\text{score}}$}}
		\Comment{ARS weight update}
		\Let{$\genparam_{\text{pema}}$}{$\genparam_{\text{pema}} + \eta \Delta \genparam_{\text{pema}}$}

	\EndWhile
	\State \Return{$\genparam_{\text{pema}}$}
	\EndFunction
\end{algorithmic}
\end{algorithm*}
In order to quantify the benefits of the chosen information-theoretic exploration, we define a pose-entropy-maximising agent (PEMA) as a comparative baseline.
PEMA is an unstructured deep exploration policy model realised through an LSTM recurrent neural network (RNN).
The chosen model boasts significant modelling capacity and, since an RNN is utilised, the ability to memorise and reason about past interactions.
What distinguishes the baseline from the method proposed in this work is mainly the lack of inductive bias in terms of the explicit probabilistic treatment of global environment properties.
And with that, the missing possibility to directly reason about gained information about them.

The LSTM policy operates in the introduced maze world, just as the DVBF-LM agent which utilises MI to perform exploration.
The parameters of the LSTM are denoted with $\genparam_{\text{pema}}$ and are to be learned.
The idea is to have the LSTM policy predict the next action $\bu_{T}$ conditioned on the history of all past observations $\bx\Ts$. 
That is:
\begin{equation*}
	\bu_T = \text{LSTM}_{\genparam_\text{pema}}(\bx\Ts).
\end{equation*}
Note that $T$ now ranges over the whole previous data set, and not over the upcoming sequence that is to be executed.
To let the LSTM learn how to explore, a training objective that corresponds to our intuition about exploration is defined.
The objective is the maximisation of the entropy of all poses in the trajectory resulting from executing the policy-selected controls.
Since entropy corresponds to diversity of the samples of a distribution, highly entropic trajectories correspond to varied positions along the trajectory, meaning more spatial regions are explored.
The baseline policy is hence coined a \emph{pose-entropy-maximising agent} (PEMA).

For convenience, the entropy metric is approximated by simply counting the number of visited cells of a grid that spans the whole maze environment.
That is:
\begin{align}\label{eq:pemaloss}
\begin{split}
	\genparam_{\text{pema}}^* &= \arg \min_{\genparam_{\text{pema}}} \loss[\text{pema}]{\bz\Ts} \\
				  &= \arg \min_{\genparam_{\text{pema}}} - \sum_{s_{ij} \in \mathcal{S}} \mathbb{I}\left( \bigvee_{t=1}^T \{ \bz_t \in s_{ij} \} \right),
\end{split}
\end{align}
where $s_{ij} \in \mathcal{S}$ denotes one of $W \times H$ discrete square cells spanning the 2D area of the whole maze in a grid.
Correspondingly, $\bigvee_{t=1}^T \{ \bz_t \in s_{ij} \}$ denotes the logical disjunction of any of the poses $\bz_t$ landing in one of those cells.
The poses $\bz\Ts$ in the above come from an \emph{actual rollout} for $T$ steps in the environment (cf. \cref{alg:pemaexec}), based on the control inputs produced by the LSTM after every step (given the previous history of observations).
For brevity, the explicit dependency of $\bz\Ts$ on $\genparam_{\text{pema}}$ is not denoted on the right hand side of \Cref{eq:pemaloss}, as obtaining $\bz\Ts$ by executing the controls in the agent's world is not a function that's differentiable in $\genparam_{\text{pema}}$.
In other words, PEMA has direct access to simulator ground truth data during training.
Still, it is assumed that such data is available for a set of maze environments, so that a comparison between the proposed method and the baseline can be conducted.
In particular, PEMA is trained on 10 mazes of average complexity and size (see \cref{fig:pemamazes}), with unlimited access to ground truth poses and observations.
The training is conducted simultaneously in all mazes, summing up the losses from each individual one to form one common objective (reward).
Since the above entropy-inspired objective is not differentiable, PEMA is trained until convergence using augmented random search (ARS) (\cite{ars}) to update the LSTM weights.
For an overview of the described optimisation, the reader is referred to \Cref{alg:pemaalg}.

\begin{figure}[t]
	\centering
	\includegraphics[width=\linewidth]{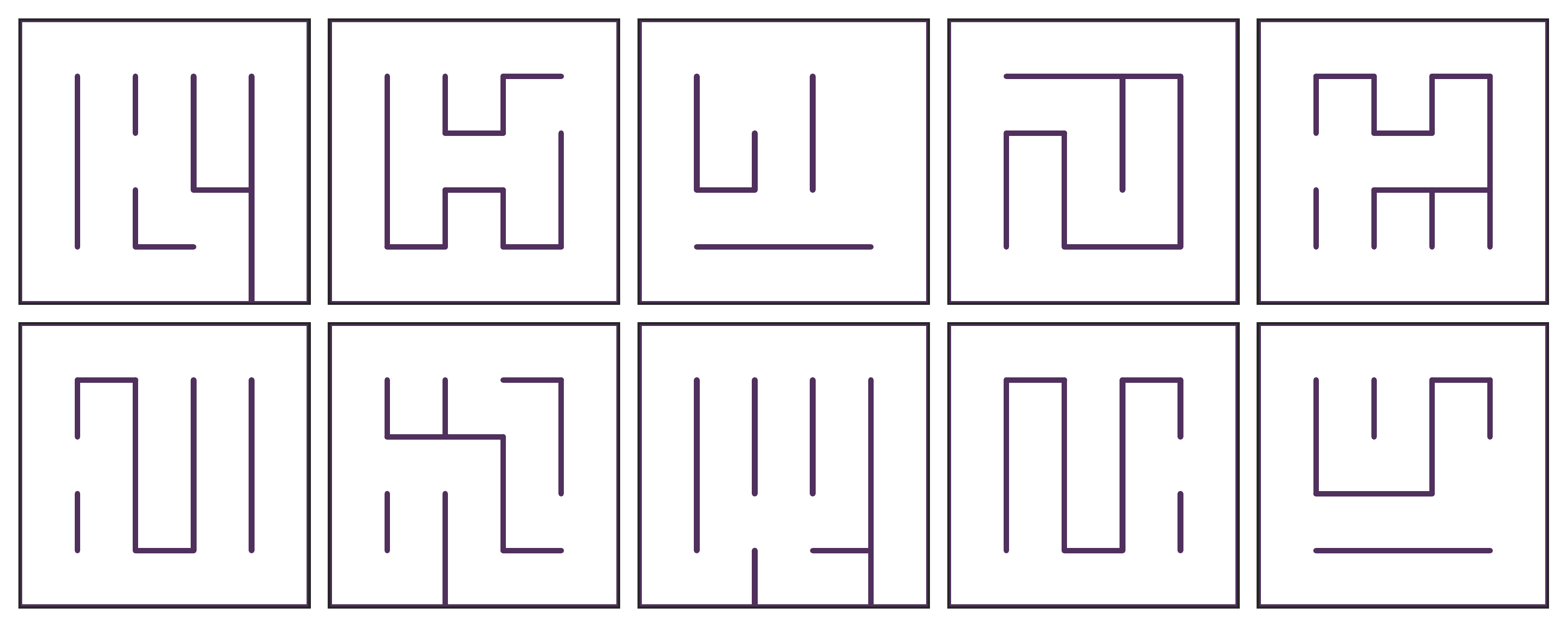}
	\caption{The ten mazes used for the training of the PEMA baseline.}\label{fig:pemamazes}
\end{figure}

\subsection{PEMA Details}
Ten different mazes of moderate complexity and average corridor width are used for the training (depicted in \Cref{fig:pemamazes}).
Every time the training objective is evaluated, the agent is placed at different random starting positions in all mazes, from where a rollout for $T = 1000$ steps following the current PEMA policy occurs.
The PEMA objective is then evaluated for the ground truth agent poses in all mazes, the rewards are summed up and an augmented random search weight update is executed.
The training procedure continues until convergence (cf.\ \Cref{alg:pemaalg}).

\subsection{Model Parameters}
\begin{itemize}
	\itemsep0em
	\item \textbf{LSTM}: The LSTM cell maps the history of laser reading observations to a single angular velocity for the current step $t$. It has 256 hidden units; a softsign activation is applied to the output states, which are fed into a feed-forward layer to produce the angular velocities. A constant position offset of $0.01$ in the direction of the agent's heading is concatenated to the produced angular velocity, and that gives the final control prediction.
	\item \textbf{Loss}: The PEMA objective is evaluated for 1000 time steps jointly in all 10 training mazes at every training iteration. $100 \times 100$ cells are used for the evaluation of the PEMA loss.
\end{itemize}

\subsection{Augmented Random Search Details}
\begin{itemize}
	\itemsep0em
	\item The number of perturbations of the parameters used for the search is set to $K_{\text{pert}} = 1$.
	\item The standard deviation of the perturbation noise is set to $\sigma_{\text{pert}} = 0.0075$.
	\item The update step length is set to $\eta = 0.001$.
\end{itemize}

}

\end{document}